\documentclass[runningheads]{llncs}
\usepackage[a4paper,left=3cm,right=3cm]{geometry}

\usepackage[T1]{fontenc}
\usepackage{graphicx}
\usepackage{verbatim}
\usepackage{amsfonts}
\usepackage{amsmath}
\usepackage{amssymb}
\usepackage{algorithm}
\usepackage{algpseudocode}
\usepackage{censor}
\usepackage{multirow}
\usepackage{booktabs}
\usepackage{array}
\usepackage{longtable}
\usepackage{tabularx}
\usepackage{threeparttable}
\usepackage{caption}
\usepackage{xurl}
\usepackage{color}
\usepackage[framemethod=default]{mdframed}
\definecolor{claimframe}{gray}{0.45}
\definecolor{claimback}{gray}{0.97}
\newmdenv[
  linewidth=0.7pt, linecolor=claimframe, backgroundcolor=claimback,
  innertopmargin=6pt, innerbottommargin=6pt, innerleftmargin=8pt, innerrightmargin=8pt,
  skipabove=9pt, skipbelow=9pt, roundcorner=2pt
]{claimbox}

\usepackage[colorlinks=true,linkcolor=blue,citecolor=blue,urlcolor=blue]{hyperref}

\begin{document}
\title{The strength of clinical evidence is recoverable from language model representations but not from their stated grades}

\author{
Soroosh Tayebi Arasteh\inst{1,2,3,4}$^{\ast}$
}

\institute{
Lab for AI in Medicine, RWTH Aachen University, Aachen, Germany \and
Department of Diagnostic and Interventional Radiology, University Hospital RWTH Aachen, Aachen, Germany \and
Department of Urology, Stanford University, Stanford, CA, USA \and
Department of Radiology, Stanford University, Stanford, CA, USA
}

\maketitle 
{\footnotesize
\noindent$^{\ast}$Correspondence to: Soroosh Tayebi Arasteh (\email{soroosh.arasteh@rwth-aachen.de})
}

\begin{abstract}
Large language models (LLMs) increasingly summarize clinical evidence, where a claim's weight depends on how strongly it is supported. Yet these models convey confidence poorly, and properties they never state, such as truth, are often readable from their activations. Whether a clinical model registers evidence strength, distinct from truth, and states it when asked is untested, and any such signal could be lexical. We compiled 45,134 clinical claims from six public sources, harmonized 20,611 into a four-level evidence grade under three independent frameworks, and tested 22 local, open-weight LLMs from several developers (0.6--70 billion parameters; general, medical, and reasoning), with lexical, truth, and cross-framework controls. A linear estimator recovered the grade in every model (median AUROC 71.8), yet decodability did not rise with scale and was weakest in reasoning models. The grade the models stated fell to chance, 25--27 percentage points below the estimator. The recoverable signal was largely lexical and did not transfer across topics or frameworks, yet it was distinct from factual truth and still flagged weakly supported claims (AUROC 69.2). Clinical LLMs thus carry an ordered evidence-strength signal they do not express, so their stated grades fail to convey a claim's support even when it is recoverable from their representations and text.
\end{abstract}

%%%%%%%%%%%%%%%%%%%%%%%%%%%%%%%%%%%%%%%%%%%%
%%%%%%%%%%%%%%%%%%%%%%%%%%%%%%%%%%%%%%%%%%%%
%%%%%%%%%%%%%%%%%%%%%%%%%%%%%%%%%%%%%%%%%%%%
%%%%%%%%%%%%%%%%%%%%%%%%%%%%%%%%%%%%%%%%%%%%

\section*{Introduction}

Large language models (LLMs) are moving from benchmark demonstrations into clinical practice, where they answer medical questions, draft clinical documentation, and summarize evidence for clinicians and patients \cite{singhal2023large,thirunavukarasu2023large,vanveen2024adapted,singhal2025expert}. In medicine the value of an assertion depends not only on whether it is correct but on how strongly it is supported: a recommendation backed by several large randomized controlled trials (RCTs) carries different weight from one resting on a single small study or on expert opinion alone. For an LLM's output to be used safely, a reader must be able to tell which of its claims are well supported and which are not.

Yet LLMs communicate confidence poorly. Their stated certainty is weakly calibrated and frequently fails to track whether an answer is correct \cite{kadavath2022language,lin2022teaching,tian2023just,xiong2024can}, and the explanations they offer can leave a reader more confident than the model's own internal estimate warrants \cite{steyvers2025calibration}. These failures are especially consequential in medicine, where current models still misjudge realistic clinical cases while expressing little hesitation \cite{hager2024evaluation,thirunavukarasu2023large}. A model that asserts a tentative finding as confidently as a settled one gives the reader no signal of evidentiary strength, the very signal that clinical judgment depends on.

Interpretability research suggests that the missing signal might nonetheless be present inside the model. Properties that LLMs do not reliably state, including the truth of a factual claim, are often linearly decodable from their hidden activations \cite{azaria2023internal,burns2023discovering,li2023inference,marks2024geometry}, and recent work argues that internal representations encode more about a model's own correctness than its outputs reveal \cite{orgad2025llms}. This raises a specific possibility for clinical evidence: a model might internally register that a claim is weakly supported while failing to convey it when prompted.

Whether this is so has not been tested, and there are reasons for caution. The strength of clinical evidence is a graded property, formalized by systems such as the Grading of Recommendations Assessment, Development and Evaluation (GRADE) framework and the recommendation grades of the US Preventive Services Task Force (USPSTF) \cite{guyatt2008grade,uspstf2018procedure}, and it is distinct from truth: well supported claims are sometimes overturned, and reversals of established practice are common enough to make the distinction concrete \cite{herrera2019reversals}. At the same time, an apparent internal signal can be shallow. Recent analyses show that hidden-state readouts often capture whether the model is drawing on memorized statistical associations instead of whether a statement is true, and that such signals are driven in part by surface lexical cues and fail to generalize across datasets \cite{cheang2026recall,orgad2025llms}. Any claim that a model encodes evidence strength must therefore be separated from lexical cues, from factual truth, and from the particular corpus on which it was measured.

Here we ask, across 22 local, open-weight LLMs spanning general-domain families, medical domain-adapted models \cite{labrak2024biomistral}, and reasoning-distilled models \cite{deepseekai2025r1,chen2024huatuogpto1} and ranging from 0.6 to 70 billion parameters, whether the evidence grade of a clinical claim is recoverable from a model's hidden states and whether the model can state it. We compile a corpus of 45{,}134 clinical claims entirely from publicly available biomedical sources, comprising automatically extracted RCT evidence \cite{marshall2020trialstreamer,lehman2019inferring,nye2018ebmnlp}, national preventive-care recommendations, and catalogued reversals of established practice \cite{herrera2019reversals}, each graded for strength of evidence under one of three independent grading frameworks. Using linear estimators, with controls for lexical and surface confounds, a balanced design that decorrelates grade from factual truth, and a leave-one-framework-out test of generalization, we characterize what these models encode and what they express.

\begin{figure}[p]
\centering
\includegraphics[width=\textwidth]{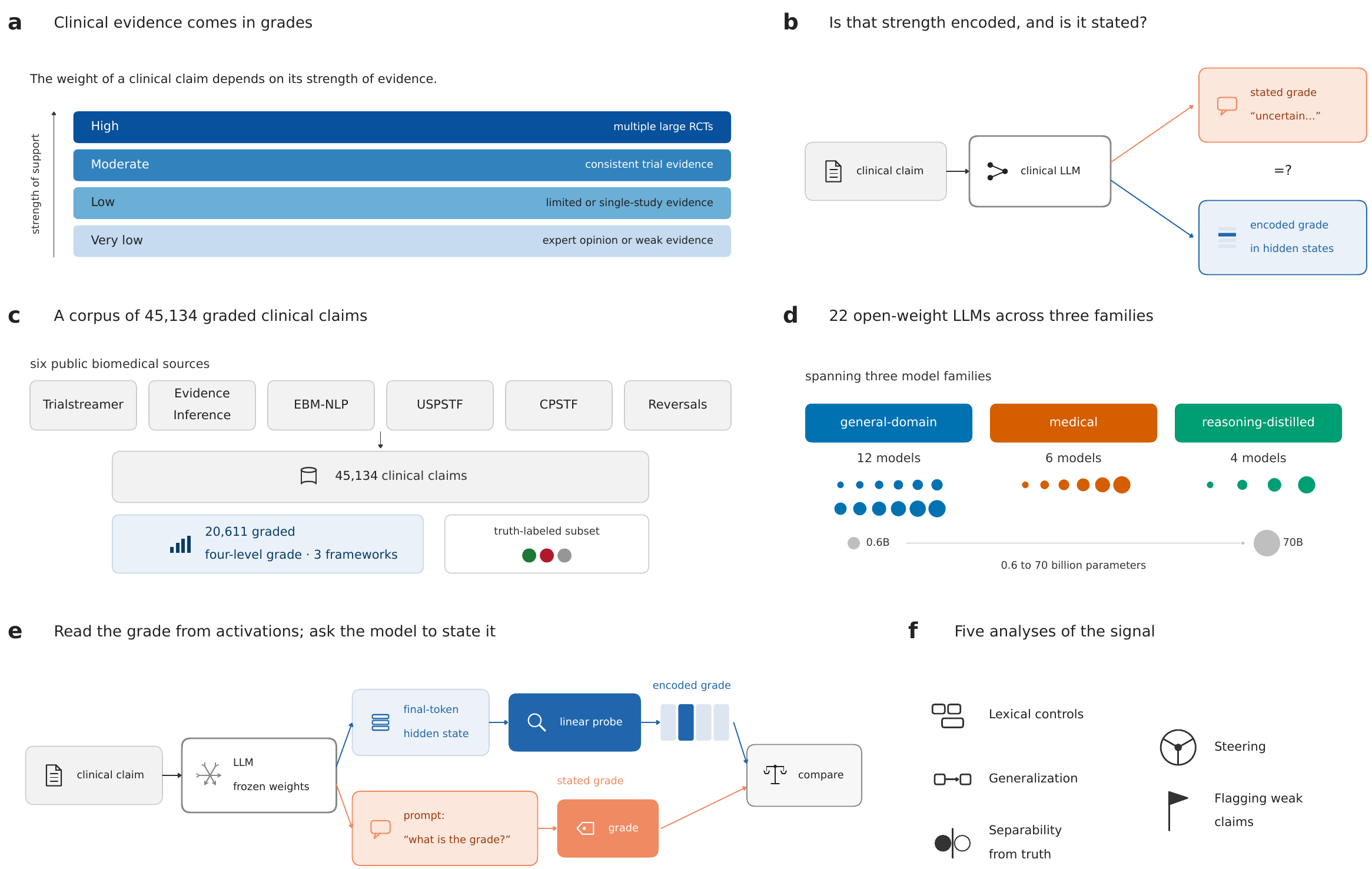}
\caption{Overview of the study. \textbf{a}, In clinical medicine, the weight of an assertion depends on how strongly it is supported by evidence, a graded property shown here on a four-level scale ordered from high to very low support. \textbf{b}, The study asks whether a language model internally registers this strength in its hidden activations and whether it states the same grade in its output, that is, whether the encoded and the stated grade agree. \textbf{c}, A corpus of $45{,}134$ clinical claims is compiled from six public biomedical sources, of which $20{,}611$ are harmonized into the four-level evidence grade under three independent grading frameworks, with a subset additionally labeled for factual truth. \textbf{d}, The corpus is used to assess $22$ open-weight language models spanning general-domain, medical domain-adapted, and reasoning-distilled families and ranging from $0.6$ to $70$ billion parameters. \textbf{e}, For each claim, the final-token hidden state of the frozen model is extracted, and a linear estimator is trained to read the evidence grade from the representation, while the model is separately prompted to state the grade, defining the central comparison between the recoverable and the stated grade. \textbf{f}, Further analyses examine whether the recoverable signal is lexical, whether it generalizes across clinical topics and grading frameworks, whether it is separable from factual truth, whether it can be steered, and whether it can flag weakly supported claims.}
\label{fig:overview}
\end{figure}

We find a consistent dissociation. Evidence grade is linearly recoverable in every model, but, contrary to the expectation that larger or more specialized models hold cleaner internal structure, it decodes no better, and often worse, in larger models and in reasoning-tuned ones. The same models cannot state the grade their activations encode: verbalized grades fall to chance, opening a large and systematic gap between what is internally recoverable and what is expressed. This recoverable signal is largely lexical and does not transfer across topics or grading frameworks, and it is separable from factual truth. Finally, although a model will not report it, the signal can be read out by an external assessor to flag weakly supported claims for review. Fig.~\ref{fig:overview} summarizes the study.

%%%%%%%%%%%%
%%%%%%%%%%%%

\section*{Methods}
\subsection*{Ethics statement}

This study used only publicly available, non-identifiable secondary data and involved no human participants and no individual patient records; it therefore required no institutional review board approval. All data sources were used in accordance with their respective licenses and terms of use; all sources are cited, and no derived corpus is redistributed. All models were accessed under, and used in compliance with, their respective licenses and acceptable-use policies; for the checkpoints whose terms required acceptance or an access request, these were completed before download.

\subsection*{Study design}

We tested whether the strength of evidence behind a clinical claim is recoverable from the internal states of frozen, open-weight LLMs and whether the same models state that strength when asked. The unit of analysis is an atomic clinical claim: a short, declarative, context-free assertion, for example that a given service benefits a given population, carrying a four-level evidence grade and, where determinable, a factual-truth label. We compiled a corpus of such claims from public biomedical sources, extracted the hidden representation of each claim from 22 LLMs spanning general-domain, medical, and reasoning-distilled families, and trained linear estimators to read the evidence grade from those representations. We then compared what the estimators recover with what the models verbalize, and ran a sequence of controls, for surface and lexical cues, for factual truth, and for generalization across topics and grading frameworks, designed to test whether any recovered signal reflects a transferable representation of evidence strength or a surface correlate. All models were used with frozen weights; no fine-tuning was performed.

\subsection*{Clinical claim corpus}

The corpus comprises $45{,}134$ unique clinical claims compiled automatically from six public sources, with no manual curation at any stage. The full construction pipeline, with per-source extraction, filtering, splitting, and all per-source and cross-tabulated statistics, is given in Supplementary Note~\ref{snote:datacuration} and Supplementary Tables~\ref{stab:sourcepipeline}, \ref{stab:gradesource}, and \ref{stab:sourcetruth}.

Three sources supply graded clinical recommendations under established frameworks. USPSTF recommendation statements, carrying grades A, B, C, D, and I, were collected from the public recommendation index ($122$ claims); topics that assign different grades to different subpopulations were split into one neutral claim per subpopulation and grade, and the claim text was built from the topic and service type alone, so that the grade never appears in the wording. Community Preventive Services Task Force (CPSTF) Community Guide findings ($224$ claims) were parsed from the official all-active-findings listing into recommended, recommended-against, and insufficient-evidence findings \cite{cpstf2024community}. Medical reversals ($395$ claims), once-standard practices later overturned by a randomized trial, were read from the eLife reversals supplement \cite{herrera2019reversals}; each carries its apparent pre-reversal grade and a false truth label, populating the otherwise empty strong-grade-but-false cell.

Three further sources supply trial-level evidence and claim diversity. Trialstreamer \cite{marshall2020trialstreamer}, an automatically maintained database of RCT reports, contributed $19{,}870$ claims built from extracted key-findings text; because its risk of bias score is strongly right-skewed, trials were graded by within-corpus quantile band instead of a fixed threshold (top band moderate, middle low, bottom very low), a relative quality tier reported as such, and single-trial findings were labeled truth-uncertain. Evidence Inference (EI) \cite{lehman2019inferring} contributed $19{,}670$ claims stating a trial's reported outcome direction; each significant-direction finding additionally emits its opposite-direction counterfactual as a verified false claim, so this source carries a large, truth-balanced true and false signal within a single writing style, while null findings are labeled truth-uncertain. EBM-NLP \cite{nye2018ebmnlp}, about $4{,}853$ RCT abstracts with manual populations, interventions, and outcomes span annotations, supplies one templated claim per abstract for topic and lexical diversity and carries neither a grade nor a truth label.

Each source was normalized to a common row schema recording the claim text, the source, the clinical topic, the evidence grade, the original framework grade, the truth label, the crossing cell, and surface covariates including hedge density, a token-frequency proxy, and claim length. Before splitting, the build removed degenerate non-claims: rows shorter than four words, rows with fewer than half alphabetic characters (statistic fragments and citations), and erratum, correction, and retraction notices. After full deduplication on claim text, the corpus contained $n=45{,}134$ claims, of which $n=20{,}611$ carry a four-level evidence grade; four of the six sources are graded and two, EI and EBM-NLP, are not (Fig.~\ref{fig:corpus}a). The full composition is given in Supplementary Table~\ref{stab:corpus}.

\subsection*{Evidence grading and factual-truth labeling}

\begin{figure*}[p]
\centering
\includegraphics[width=0.95\textwidth]{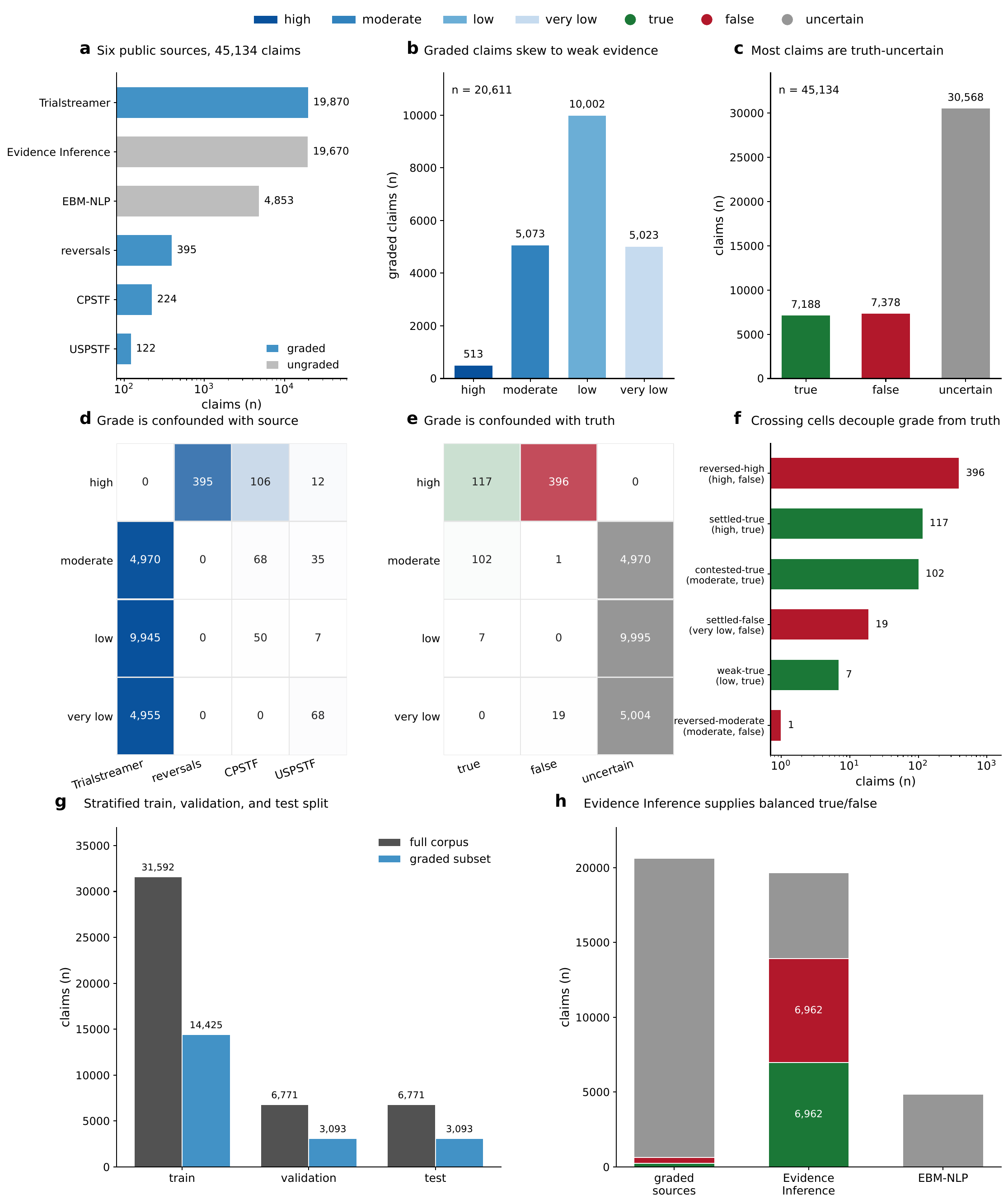}
\caption{Composition of the clinical evidence-grade claim corpus ($n=45{,}134$ claims, of which $n=20{,}611$ carry a four-level evidence grade). \textbf{a}, Number of claims from each of the six public sources, separated into graded and ungraded sources, on a logarithmic scale. \textbf{b}, Distribution of the four evidence grades among the $n=20{,}611$ graded claims. \textbf{c}, Distribution of factual-truth labels across the full corpus. \textbf{d}, Grade-by-source composition of the graded subset; each cell gives the number of claims and is shaded by its within-grade proportion, and USPSTF is the only source spanning all four grades. \textbf{e}, Grade-by-truth contingency of the graded subset, with cell counts; the two labels are strongly associated (Cramer's $V=0.614$~\cite{cramer1946mathematical}). \textbf{f}, Number of claims in each of the six crossing cells, the rare cells in which grade and truth are decoupled, on a logarithmic scale and colored by truth label; these cells are the basis of the balanced grade-by-truth analysis. \textbf{g}, Sizes of the crossing-cell-stratified $70/15/15$ training, validation, and test partitions, for the full corpus and for the graded subset. \textbf{h}, Factual-truth composition by source group, showing that EI contributes the matched true and false claims ($n=6{,}962$ each) that decouple truth from writing style. Evidence grades and truth labels are colored as in the legend.}
\label{fig:corpus}
\end{figure*}

Each framework's native grade was harmonized to a common four-level ordinal scale (high, moderate, low, and very low). USPSTF grades A, B, C, and D or I map to high, moderate, low, and very low; CPSTF recommended and recommended-against findings at strong strength map to high and at sufficient or unstated strength to moderate, and insufficient-evidence findings to low; Trialstreamer's top, middle, and bottom rigor bands map to moderate, low, and very low; and reversals carry their apparent pre-reversal grade, high or, where annotated, moderate. The full mapping is given in Supplementary Table~\ref{stab:harmonization}, and a sensitivity analysis on the mapping, substituting an absolute-threshold rule for the Trialstreamer quantile bands, is reported alongside the main decodability result. The graded subset ($n=20{,}611$) comprises $513$ high, $5{,}073$ moderate, $10{,}002$ low, and $5{,}023$ very-low claims (Fig.~\ref{fig:corpus}b).

A factual-truth label (true, false, or truth-uncertain) was assigned independently of grade. A claim is true when it reflects current best evidence (USPSTF grades A, B, and C, and CPSTF recommendations), false when it states a practice a trial has overturned (the $395$ reversals, USPSTF grade D, and the two CPSTF recommended-against findings), and truth-uncertain when the evidence is insufficient or the finding is a single trial or a null result (USPSTF grade I, and most Trialstreamer and EI items). Across the corpus, $7{,}188$ claims are labeled true, $7{,}378$ false, and $30{,}568$ truth-uncertain (Fig.~\ref{fig:corpus}c).

In the graded subset, grade and truth are associated: strong-grade claims carry definite truth labels while weak-grade claims are predominantly truth-uncertain (Fig.~\ref{fig:corpus}e). Within this association, a small set of crossing cells decouples the two (Fig.~\ref{fig:corpus}f). Reversed-high ($n=396$) and reversed-moderate ($n=1$) claims are strong-grade but false; settled-true ($n=117$) and contested-true ($n=102$) claims are strong-grade and true; and weak-true ($n=7$, USPSTF grade C) and settled-false ($n=19$, USPSTF grade D) claims populate the otherwise sparse low and very-low true and false cells. These cells are what make grade separable from truth and are the basis of the balanced grade-by-truth analysis described below.

Grade is strongly confounded with source and writing style in the assembled corpus: very-low claims are almost all from Trialstreamer, strong-grade claims are reversals and CPSTF findings, and within the high grade every true claim is a CPSTF finding while every false claim is a reversal (Fig.~\ref{fig:corpus}d). Two design elements guard the central analyses against a classifier that reads source. First, EI provides a within-source, single-template, truth-balanced set on which a truth direction can be learned in isolation from writing style (Fig.~\ref{fig:corpus}h). Second, USPSTF is the only source spanning the full grade range within one rubric and one style, which both breaks the grade-source confound and serves as the held-out target for the cross-framework transfer analysis. The small grade-by-truth cells that USPSTF contributes (seven weak-true and nineteen settled-false claims) are reported pooled or with explicit bootstrap 95\% confidence intervals (CIs).

Claims were partitioned into training ($n=31{,}592$), validation ($n=6{,}771$), and test ($n=6{,}771$) sets by a claim-level, crossing-cell-stratified 70/15/15 split, so that each populated crossing cell appears in all three partitions; the single reversed-moderate claim necessarily falls in one partition. The graded subset splits into $14{,}425$ training, $3{,}093$ validation, and $3{,}093$ test claims (Fig.~\ref{fig:corpus}g). Claim text is unique across the corpus, with no text leakage and no grade or truth contradiction between partitions.

%%%%%%%%%%%%
%%%%%%%%%%%%

\subsection*{Language models}

We tested 22 local, open-weight, instruction-tuned LLMs chosen to vary model scale, domain specialization, and reasoning training while keeping families comparable; all are listed with their checkpoints and probing depths in Supplementary Table~\ref{stab:models}. Twelve are general-domain models: the Qwen3 dense ladder at six sizes from 0.6 to 32 billion parameters \cite{qwen2025qwen3}, which provides a single-recipe scale axis, together with Phi-3.5-mini \cite{abdin2024phi3}, Mistral-7B \cite{jiang2023mistral}, Llama-3.1-8B and Llama-3.3-70B \cite{grattafiori2024llama3}, and Gemma-2-9B and Gemma-2-27B \cite{gemmateam2024gemma2}. Six are medical domain-adapted models spanning the same size range: BioMistral-7B \cite{labrak2024biomistral}, Meditron-7B \cite{chen2023meditron}, OpenBioLLM-8B \cite{pal2024openbiollm}, Med42-8B and Med42-70B \cite{christophe2024med42}, and MedGemma-27B \cite{sellergren2025medgemma}. Four are reasoning-distilled models: three DeepSeek-R1 distillations, into Qwen-7B, Llama-8B, and Qwen-32B \cite{deepseekai2025r1}, and the medical reasoning model HuatuoGPT-o1-8B \cite{chen2024huatuogpto1}. Several medical and reasoning models share a base checkpoint with a general model in the panel; OpenBioLLM, Med42, and the DeepSeek-R1 distillations are derived from Llama or Qwen models, which lets the medical-versus-base and reasoning-versus-base contrasts be read at matched scale.

All checkpoints were obtained from the Hugging Face Hub and run locally with frozen weights and no adaptation. Models were loaded in half precision (float16), except the six models of 27 billion parameters and above (the two 27-billion, the two 32-billion, and the two 70-billion-parameter models), which were loaded in 4-bit quantization to fit available memory. The forward pass used a single, deterministic encoding of each claim with caching disabled, identical across the panel so that every model saw the same inputs and comparable numerics, and the activation at the final claim token was taken from the position given by the attention mask; model-specific loading constraints, such as the attention and cache settings required by Phi-3.5 and the Gemma-2 family and the context limit of the Meditron checkpoints, are recorded in the released configuration. Exact checkpoints, parameter counts, and the per-model probing layer are given in Supplementary Table~\ref{stab:models}.

\subsection*{Hidden-state representation extraction}

For each claim and each model, we extracted the residual stream activation vector at the final claim token from every transformer layer $l \in \{0,\dots,L\}$, where layer $0$ is the embedding output and $L$ is the final layer. These per-layer vectors are the features for all decoding analyses. For the geometric analyses that compare classifier directions, vectors were $L_2$-normalized. Activations were cached per model in aligned claim order. A small number of non-finite activation cells, a half-precision overflow artifact affecting on the order of $10^{-6}$ of cells in a few models, were repaired once by per-layer column-mean imputation with a logged count; any model exceeding $1\%$ non-finite cells would have been re-extracted, and none reached that level.

\subsection*{Linear probing of evidence grade}

We read the evidence grade from the hidden states with an $L_2$-regularized multinomial logistic estimator (inverse regularization strength $C=1.0$, standardized features) fit to convergence on the training split; all reported grade-decoding metrics use this estimator. For the four-level grade we report both a one-versus-rest performance per grade level and an ordinal readout: mapping the grades to ranks $r_k\in\{1,2,3,4\}$, the expected ordinal score for a claim is
\begin{equation}
s(\mathbf{h})=\sum_{k=1}^{4} r_k\, p_k(\mathbf{h}),
\label{eq:ordinal}
\end{equation}
where $p_k(\mathbf{h})$ is the estimated probability of grade $k$; we summarize the agreement of $s$ with the gold grade by its Spearman correlation and by pairwise concordance. This expected ordinal score is also the per-claim grade readout used for the cross-model agreement analysis below.

The geometric analyses that compare directions (the separability of grade and truth, below) use two optimization-free linear readouts in place of the trained estimator. A difference-in-means (mass-mean) direction, the construction prior work finds most causally implicated in model behavior \cite{li2023inference,marks2024geometry}, defines a unit direction for a binary contrast between claim sets $\mathcal{P}$ and $\mathcal{N}$,
\begin{equation}
\mathbf{w}=\frac{1}{|\mathcal{P}|}\sum_{i\in\mathcal{P}}\mathbf{h}_i-\frac{1}{|\mathcal{N}|}\sum_{j\in\mathcal{N}}\mathbf{h}_j,
\label{eq:massmean}
\end{equation}
where $\mathbf{h}_i$ is the hidden state of claim $i$; it provides the truth direction in the separability analysis, fit on true vs false claims. A ridge regression of the ordinal grade onto the activations defines the one-dimensional grade direction used in that analysis and for the activation steering below. Both directions are $L_2$-normalized to unit length.

The estimators were fit on the training split, the per-model peak layer was selected on the validation split, and all reported metrics are computed on the held-out test split. Per model, we report the macro one-versus-rest area under the receiver operating characteristic curve (AUROC), the per-class AUROC and area under the precision-recall curve (AUPRC), the four-way classification accuracy, and the ordinal Spearman correlation, together with the full layerwise decodability curve and the peak-layer value. Decodability was compared against a null in which grade labels are shuffled before fitting. To test how decodability depends on model properties, we fit the peak-layer decodability against the logarithm of parameter count by ordinary least squares (OLS)~\cite{seber2003linear}, separately over the Qwen3 ladder and over the full panel; compared each medical model with its base model and the reasoning-distilled models with non-reasoning models; and measured whether different models encode a shared per-claim grade ordering through the pairwise Spearman agreement of their grade readouts.

%%%%%%%%%%%%
\subsection*{Verbalized grade elicitation}

To test whether a model states the evidence grade it internally encodes, we elicited a grade from each model directly, independently of any external estimator. For every graded test claim, the model was prompted with a fixed instruction to rate the strength of evidence and to answer with exactly one of the four grades (high, moderate, low, or very low), decoded greedily, under two regimes: zero-shot, with the instruction and the four options but no examples, and few-shot, with two labeled exemplars per grade (eight in total, balanced across the four grades and sampled deterministically from the training split) prepended to the same instruction. The stated grade was parsed from the model's response by matching the grade vocabulary, with ``very low'' matched ahead of ``low''; responses from which no grade could be recovered were counted as errors. A verbalized readout was kept only when it formed a usable signal: it had to yield a parseable grade for at least half of the test claims and had to be non-degenerate, with no single grade exceeding $80\%$ of its parsed responses. Six of the 22 models failed one of these conditions under zero-shot prompting, either collapsing onto a single grade or not following the answer format, and were excluded; the gap is therefore measured only on the models that returned a usable graded response, and is, if anything, conservative. For each retained model we computed the verbalized four-way grading accuracy and the ordinal Spearman correlation of the stated grade with the gold grade, and compared both to the internal logistic estimator evaluated on the same claims. The internal-versus-verbalized comparison is paired at the claim level: we report the difference in four-way accuracy, the behavioral gap, with a paired bootstrap~\cite{efron1993introduction}, and McNemar's test~\cite{mcnemar1947note} on the paired correct and incorrect outcomes, distinguishing among discordant claims those the estimator alone classified correctly from those the model alone stated correctly.

\subsection*{Lexical and surface-confound controls}

Because evidence grade could, in principle, be read from surface features of the claim text instead of from a representation of evidence strength, we ran three families of control.

First, we compared the hidden-state estimator against surface baselines fit to the same claims: a term frequency-inverse document frequency (TF-IDF) \cite{tfidfmain} bag-of-words \cite{joachims1998text} logistic classifier over word and character n-grams, and a classifier on the measured surface covariates (claim length, hedge density, and a token-frequency proxy). We also re-fit the grade estimator within a single source, where source and writing style are held constant, and computed the partial Spearman correlation between the hidden grade-score and the gold grade after residualizing out those surface covariates, requiring decodability to persist in each case.

Second, at each model's peak layer we fit three classifiers on the same split, a lexical-only (TF-IDF) classifier, a hidden-only estimator, and a combined estimator that augments the hidden features with the lexical grade-score, and measured both the combined-minus-lexical macro-AUROC increment, with a paired bootstrap over test claims, and the partial Spearman correlation between the hidden grade-score and the gold grade after partialling out the lexical grade-score. The TF-IDF classifier was fit once and cached, since all models share identical claim text.

Third, we ran a matched-text test that removes the lexical advantage by construction. At each model's peak layer the hidden-state estimator and the TF-IDF classifier each assign every test claim an ordinal grade-score. We then formed pairs of test claims with different gold grades, sampling up to $20{,}000$ pairs, computed the TF-IDF cosine similarity of each pair, and stratified the pairs by similarity into terciles and a strict near-duplicate stratum (cosine at least $0.8$). Within each stratum we measured the pairwise ordering accuracy of each score, the fraction of pairs whose score difference has the same sign as the gold grade difference (ties counted as one half), and the paired bootstrap difference between the hidden and lexical ordering accuracies. On lexically near-identical pairs a bag-of-words score is at chance by construction, so any advantage of the hidden score in that stratum would be evidence of grade information beyond surface words.

\subsection*{Generalization across topics and grading frameworks}

We tested whether an estimator trained on one part of the corpus reads evidence grade elsewhere, along three axes. For cross-topic generalization, we fit the grade estimator on a set of clinical topics and evaluated it on held-out topics, reporting both the pooled transfer and the per-topic transfer. For cross-framework generalization, we exploited the three independent grading rubrics that label claims on the common scale, the USPSTF clinical-prevention grades, the CPSTF community-intervention strengths, and the Trialstreamer trial-level rigor bands: leaving one rubric out, we trained the estimator on the other two and evaluated it on the held-out rubric, against a per-source permutation null and a check of whether the training and held-out grade directions align more than chance. To ask whether transfer improves once source identity is removed, we additionally fit a logistic source classifier on the training sources only, projected the held-out representations onto the complement of its top eight coefficient directions (the source-identifying subspace), and repeated the leave-one-rubric-out transfer on the residual representations, comparing the result both to the permutation null and to plain transfer. Finally, for the reversal claims, which carry a pre-reversal and a post-reversal label, we tested which evidence state the estimator reads relative to each model's training cutoff, a check that the estimator tracks the encoded evidence state.

%%%%%%%%%%%%
\subsection*{Separability of evidence grade and factual truth}

Evidence grade and factual truth are correlated in the natural corpus, so an estimator that appears to read one could be reading the other. To test their separability cleanly we built a balanced grade-by-truth grid on which the two are statistically independent by construction. The grid combines corpus claims that carry both a grade and a truth label with EI claims, whose truth is the trial's reported direction and whose counterfactuals supply matched false claims of the same form; each EI claim was given a heuristic grade, a significant finding low and a null finding very low, stated as a heuristic. Within each grade the two truth cells were equalized to their shared size (capped at $2{,}000$ claims per cell), and a grade was retained only if both of its truth cells reached at least $50$ claims. On the real data this retained the low grade ($2{,}000$ claims per truth cell) and the high grade ($117$ per cell) and dropped the moderate and very-low grades, yielding a grid on which grade and truth are essentially uncorrelated. Because the truth direction must not become a source detector, it was learned on EI alone, a within-source, single-template, truth-balanced set.

On this grid, we ran the dissociation tests. The key test is truth-within-grade: holding the grade fixed, we asked whether an estimator decodes truth, fit separately per retained grade. We also decoded grade across the grid, measured how well the grade direction predicts truth and the truth direction predicts grade, and, where at least two grades were present, computed the angle between the grade and truth-estimator directions,
\begin{equation}
\cos\theta=\frac{\langle\mathbf{w}_g,\mathbf{w}_t\rangle}{\lVert\mathbf{w}_g\rVert\,\lVert\mathbf{w}_t\rVert},\qquad \theta=\arccos\lvert\cos\theta\rvert,
\label{eq:cosangle}
\end{equation}
where $\mathbf{w}_g$ is the ridge grade direction, $\mathbf{w}_t$ the difference-in-means truth direction, and $\theta$ near $90^{\circ}$ indicates near-orthogonal axes; both the cosine and the principal angle were obtained from a bootstrap that refits both directions on each resample. We report the dissociation only at the grades the balanced data support. Within the high grade the truth contrast coincides with a source contrast, CPSTF findings vs reversals, and is reported as a source artifact, so the load-bearing result is truth-within-low. As a foil, we ran the same direction and cross-prediction analyses on the natural, confounded corpus, where they cannot be read as a dissociation; the contrast between the two motivates the balanced design.

\subsection*{Activation steering along the grade direction}

To test whether the grade direction is functional during generation, not merely decodable, we steered along it and measured the effect on the model's expressed certainty. At a model's peak layer, we added the unit-norm ridge grade direction $\hat{\mathbf{w}}_g$ to the residual stream at the final claim token during generation, scaled by the mean residual norm at that layer,
\begin{equation}
\mathbf{h}'_l=\mathbf{h}_l+\alpha\,\bar{h}_l\,\hat{\mathbf{w}}_g,
\label{eq:steer}
\end{equation}
where $\mathbf{h}_l$ is the residual-stream state at the inject layer, $\bar{h}_l$ is the mean residual-stream norm over claims at that layer (so $\alpha$ is a fraction of that norm), and $\alpha$ is a signed steering coefficient. We swept $\alpha$ in both directions and measured the model's stated certainty from its certainty-token probabilities, summarizing the response as the slope of stated certainty against $\alpha$ over the range in which generations remain coherent. Coherence was tracked by the perplexity and the repetition fraction of the steered continuations under the unsteered model, and the slope was compared with that obtained when steering along a matched random direction, averaged over three random draws. This analysis was run on six of the seven models selected for steering; MedGemma-27B was excluded because its decoder layers could not be reliably located for injection. We draw no causal conclusion from the result.

\subsection*{Evidence-strength triage}

As a deployment-facing demonstration, we trained, at each model's grade-decoding peak layer, a binary linear estimator that separates weak-evidence claims (the lower grades) from stronger ones, a single-forward-pass detector of weakly supported claims that requires no retrieval or external evidence. The estimator was fit on the training split and evaluated in distribution on the held-out test split. For each model we report the flag AUROC, AUPRC, accuracy, and Brier score~\cite{brier1950verification}, together with the flag rate, precision, recall, and specificity at a set of target flag rates.

\subsection*{Statistical analysis}

\paragraph{Reporting conventions.} Performance metrics (AUROC, AUPRC, accuracy, agreement rates, and the internal-minus-verbalized gap) are reported as percentages, as a bootstrap mean followed by its standard deviation (SD) and a 95\% confidence interval (CI) shown in brackets; for brevity throughout the paper, the percent sign is omitted and the interval is printed as a bracketed range. A difference between two metrics is itself a metric, reported in the same form and additionally carrying the p-value of its significance test. Statistical measures (p-values, Spearman and partial correlations, Cramer's V, cosine, principal angle, OLS slope, and cross-model agreement) and Brier score are reported as raw values to three decimal places. A p-value from an exact test below $0.001$ is given in scientific notation.

\paragraph{Resampling and tests.} Significance tests are two-sided, except the directional bootstrap test of whether adding the hidden features improves on the lexical grade classifier, which is one-sided. Confidence intervals and resampled p-values use $10{,}000$ bootstrap resamples~\cite{efron1993introduction}, except the combined-minus-lexical AUROC increment ($2{,}000$ resamples) and the direction-geometry cosine and angle ($1{,}000$ resamples, whose reported point estimates do not depend on the resample count); permutation tests use $1{,}000$ permutations; all inferential statistics use a fixed random seed. The resampling unit is the claim for the decodability and separability analyses and the grade level or clinical specialty for the corresponding regressions and subgroup comparisons. We fixed a single test for each situation and held it constant throughout: a percentile bootstrap for the mean of a metric over a set of units; a paired bootstrap~\cite{efron1993introduction} for the difference between two conditions on the same claims; a Spearman correlation with its analytic p-value for the association of two aligned vectors, and a residualized Spearman for a partial correlation controlling a covariate; an OLS fit~\cite{seber2003linear} with its slope p-value and $R^2$ for the scaling and steering fits; a permutation test for two- or $k$-group mean comparisons with small cells; a bootstrap that refits both directions for the cosine and principal angle; and Cram\'er's V~\cite{cramer1946mathematical} with a chi-square p-value for the grade-by-truth association. McNemar's test~\cite{mcnemar1947note} was used for the paired comparison of internal and verbalized correctness.

\paragraph{Multiplicity and software.} Within each family of tests, defined by the experiment and the analysis, p-values were adjusted for multiple comparisons by the Benjamini-Hochberg false discovery rate (FDR) procedure~\cite{benjamini1995controlling}, and both raw and adjusted values are reported. Significance was assessed at the FDR-adjusted threshold of $\alpha=0.05$. Tests in families comprising a single comparison carry an adjusted p-value equal to their raw p-value; those are labeled $p$ and not $p_{\mathrm{FDR}}$, and every $p_{\mathrm{FDR}}$ label marks a multi-comparison family. Descriptive and inferential quantities are kept distinct throughout. All analyses used a single environment (Python~3.11, with packages: NumPy~1.26, SciPy~1.17, pandas~3.0, and scikit-learn~1.8).

\section*{Results}

%%%%%%%%%%%%%%%%%%%%%%%%%%%%%%%%%%%%%%%%%%%%%%%%%%%%%%%
\subsection*{Evidence grade is linearly recoverable from hidden states, and decodability inverts with model scale}

\begin{figure*}[p]
\centering
\includegraphics[width=0.97\textwidth]{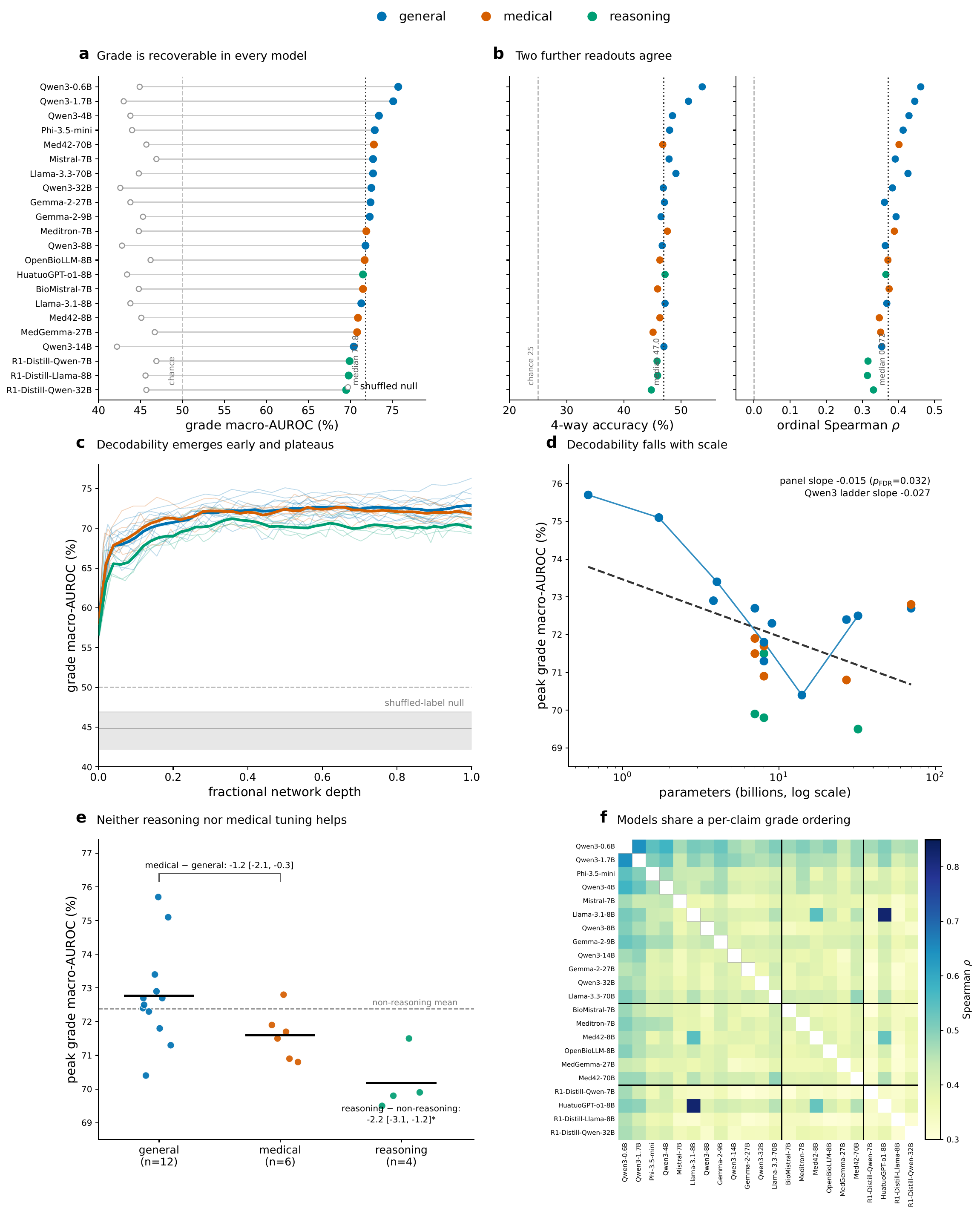}
\caption{Decodability of the four-level evidence grade from hidden states across the 22-model panel, on the graded held-out test set ($n=3{,}093$ claims). \textbf{a}, Per-model peak-layer macro one-versus-rest grade-decoding AUROC (filled markers) and each model's shuffled-label null (open markers), with models ordered by AUROC and the dashed line marking the panel median. \textbf{b}, Per-model four-way grade accuracy (left) and the Spearman correlation between the estimator's scalar projection and grade (right), in the same model order, with chance and zero reference lines. \textbf{c}, Layerwise macro grade AUROC against fractional network depth, one thin line per model and a bold mean per family, with the shuffled-label null band. \textbf{d}, Peak AUROC against parameter count on a logarithmic axis, with the Qwen3 size ladder connected and ordinary-least-squares fits for the ladder and for the full panel. \textbf{e}, Per-model peak AUROC grouped by model family, with the family mean of each group and the reasoning-minus-non-reasoning and medical-minus-general differences (95\% CIs) annotated. \textbf{f}, Pairwise Spearman correlation of the per-claim grade readout across all 231 model pairs, with models grouped by family. Markers are colored by model family as in the legend.}
\label{fig:decodability}
\end{figure*}

A linear estimator trained on the residual stream recovers the four-level evidence grade in every model, well above a shuffled-label null. Across the 22 models the peak-layer macro one-versus-rest AUROC has a median of 71.8 (range 69.5 to 75.7), against a null near chance (median 44.8), exceeding it by $27.3 \pm 0.5$ [26.4, 28.3] on average (Fig.~\ref{fig:decodability}a). Four-way grade accuracy has a median of 47.0 (range 44.8 to 53.7), well above the 25.0 expected by chance, and a scalar projection along the learned direction orders held-out claims by grade monotonically (Spearman correlation, median 0.372, range 0.314 to 0.462, significant after correction in all 22 models; Fig.~\ref{fig:decodability}b). The signal is not localized: decoding rises from the embedding layer to roughly the lower third of the network and then holds flat through the upper layers (Fig.~\ref{fig:decodability}c), so the precise peak layer (Table~\ref{tab:decodability}) is not load-bearing. Per-model peak values with their 95\% CIs are listed in Table~\ref{tab:decodability}. Evidence grade is therefore present and recoverable, and stable across architectures.

Decodability does not improve with model capacity, and if anything it weakens. On the controlled Qwen3 ladder, where the tokenizer and training recipe are held fixed across six sizes, macro AUROC falls from 75.7 at 0.6B parameters to 70.4 at 14B, and the ordinary least-squares regression of peak AUROC on log-parameters has a significantly negative slope ($-0.027$, $p_{\mathrm{FDR}}=0.032$); the same negative slope holds across the full panel ($-0.015$, $p_{\mathrm{FDR}}=0.032$) (Fig.~\ref{fig:decodability}d). The smallest model is the single best decoder. Reasoning-distilled models are the weakest, decoding grade $2.2$ below their non-reasoning counterparts ($-2.2 \pm 0.5$ [$-3.1$, $-1.2$], permutation test, $p_{\mathrm{FDR}}=0.018$), whereas medical post-training has no reliable effect ($-1.2 \pm 0.5$ [$-2.1$, $-0.3$], $p_{\mathrm{FDR}}=0.067$) (Fig.~\ref{fig:decodability}e). A signal read out more accurately from smaller and less specialized models does not behave like a competence that scale or domain adaptation sharpens.

Although each estimator is fit in its own model's representation space, the per-claim grade readouts agree across models. The rank correlation of grade scores between model pairs has a median of 0.393 (range 0.299 to 0.823) and is significant after correction for all 231 pairs (Fig.~\ref{fig:decodability}f), so architectures, scales, and training regimes that differ widely still recover a common ordering of which claims rest on stronger or weaker evidence. Whether this internally available signal is of any use to a reader, however, depends on the model being able to state it.

%%%%%%%%%%%%
%%%%%%%%%%%%

\begin{table}[p]
\centering
\caption{Per-model decodability of the four-level evidence grade from hidden states, on the graded held-out test set ($n=3{,}093$ claims; high $77$, low $1{,}501$, moderate $761$, very low $754$). Models are grouped by family and ordered by parameter count within each group; HuatuoGPT-o1-8B is a medical reasoning model and is listed with the reasoning-distilled group. Params, parameter count in billions. Peak layer is the residual-stream layer selected on the validation split, with its fractional network depth (layer divided by number of layers) in parentheses. Grade macro-AUROC is the macro one-versus-rest area under the receiver operating characteristic curve, and Grade accuracy is four-way classification accuracy; both are percentages, reported as mean $\pm$ standard deviation [95\% CI] over $10{,}000$ bootstrap resamples of the test claims.}
\label{tab:decodability}
\footnotesize
\setlength{\tabcolsep}{5pt}
\renewcommand{\arraystretch}{1.15}
\begin{tabular}{@{}lcccc@{}}
\toprule
Model & Params (B) & Peak layer (depth) & Grade macro-AUROC & Grade accuracy \\
\midrule
\multicolumn{5}{@{}l}{\textit{General-domain}} \\
Qwen3-0.6B & 0.6 & 27~(0.96) & 75.7 $\pm$ 0.7 [74.4, 77.0] & 53.7 $\pm$ 0.9 [52.0, 55.4] \\
Qwen3-1.7B & 1.7 & 28~(1.00) & 75.1 $\pm$ 0.6 [74.0, 76.3] & 51.3 $\pm$ 0.9 [49.6, 53.1] \\
Qwen3-4B & 4 & 13~(0.36) & 73.4 $\pm$ 0.6 [72.2, 74.6] & 48.5 $\pm$ 0.9 [46.7, 50.2] \\
Qwen3-8B & 8 & 20~(0.56) & 71.8 $\pm$ 0.7 [70.4, 73.1] & 46.7 $\pm$ 0.9 [44.9, 48.4] \\
Qwen3-14B & 14 & 37~(0.93) & 70.4 $\pm$ 0.8 [68.8, 72.0] & 47.0 $\pm$ 0.9 [45.3, 48.8] \\
Qwen3-32B & 32 & 55~(0.86) & 72.5 $\pm$ 0.7 [71.2, 73.8] & 46.9 $\pm$ 0.9 [45.1, 48.7] \\
Phi-3.5-mini & 3.8 & 20~(0.63) & 72.9 $\pm$ 0.6 [71.7, 74.1] & 48.0 $\pm$ 0.9 [46.3, 49.8] \\
Mistral-7B-v0.3 & 7 & 16~(0.50) & 72.7 $\pm$ 0.6 [71.4, 73.9] & 47.9 $\pm$ 0.9 [46.2, 49.8] \\
Llama-3.1-8B & 8 & 28~(0.88) & 71.3 $\pm$ 0.8 [69.6, 72.8] & 47.2 $\pm$ 0.9 [45.4, 48.9] \\
Gemma-2-9B & 9 & 27~(0.64) & 72.3 $\pm$ 0.6 [71.1, 73.4] & 46.5 $\pm$ 0.9 [44.7, 48.2] \\
Gemma-2-27B & 27 & 45~(0.98) & 72.4 $\pm$ 0.6 [71.2, 73.5] & 47.1 $\pm$ 0.9 [45.3, 48.8] \\
Llama-3.3-70B & 70 & 58~(0.73) & 72.7 $\pm$ 0.7 [71.2, 74.1] & 49.1 $\pm$ 0.9 [47.3, 50.9] \\
\midrule
\multicolumn{5}{@{}l}{\textit{Medical domain-adapted}} \\
BioMistral-7B & 7 & 31~(0.97) & 71.5 $\pm$ 0.8 [69.9, 72.9] & 45.9 $\pm$ 0.9 [44.2, 47.7] \\
Meditron-7B & 7 & 31~(0.97) & 71.9 $\pm$ 0.7 [70.4, 73.2] & 47.6 $\pm$ 0.9 [45.8, 49.3] \\
OpenBioLLM-8B & 8 & 20~(0.63) & 71.7 $\pm$ 0.8 [70.1, 73.1] & 46.3 $\pm$ 0.9 [44.5, 48.0] \\
Med42-8B & 8 & 29~(0.91) & 70.9 $\pm$ 0.8 [69.4, 72.5] & 46.3 $\pm$ 0.9 [44.5, 48.0] \\
MedGemma-27B & 27 & 34~(0.55) & 70.8 $\pm$ 0.6 [69.6, 72.0] & 45.1 $\pm$ 0.9 [43.3, 46.8] \\
Med42-70B & 70 & 76~(0.95) & 72.8 $\pm$ 0.6 [71.6, 74.0] & 46.8 $\pm$ 0.9 [45.0, 48.5] \\
\midrule
\multicolumn{5}{@{}l}{\textit{Reasoning-distilled}} \\
R1-Distill-Qwen-7B & 7 & 22~(0.79) & 69.9 $\pm$ 0.7 [68.6, 71.2] & 45.8 $\pm$ 0.9 [44.0, 47.6] \\
R1-Distill-Llama-8B & 8 & 22~(0.69) & 69.8 $\pm$ 0.8 [68.2, 71.2] & 45.9 $\pm$ 0.9 [44.2, 47.7] \\
R1-Distill-Qwen-32B & 32 & 64~(1.00) & 69.5 $\pm$ 0.8 [67.8, 71.0] & 44.8 $\pm$ 0.9 [43.0, 46.5] \\
HuatuoGPT-o1-8B & 8 & 28~(0.88) & 71.5 $\pm$ 0.8 [69.8, 73.0] & 47.2 $\pm$ 0.9 [45.4, 49.0] \\
\bottomrule
\end{tabular}
\end{table}

%%%%%%%%%%%%%%%%%%%%%%%%%%%%%%%%%%%%%%%%%%
%%%%%%%%%%%%%%%%%%%%%%%%%%%%%%%%%%%%%%%%%%

\subsection*{Models cannot state the evidence grade their activations encode}

\begin{figure*}[p]
\centering
\includegraphics[width=0.97\textwidth]{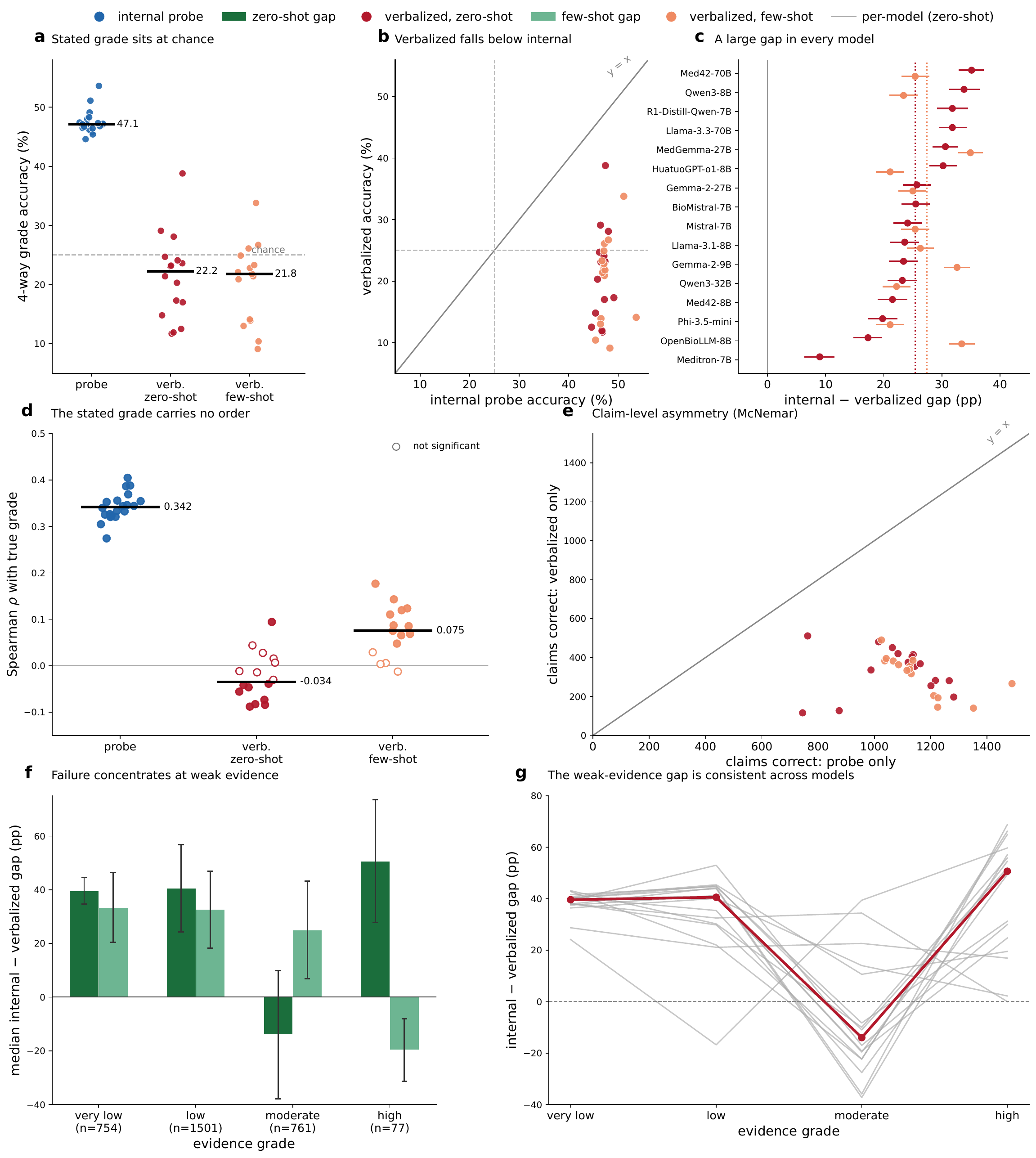}
\caption{Internally recoverable evidence grade compared with the grade the models state when asked, on the graded held-out test set, over the models that returned a parseable verbalized grade (16 zero-shot, 15 few-shot). \textbf{a}, Four-way grade accuracy for the linear estimator and for the verbalized grade under zero-shot and few-shot prompting, one dot per model with group medians and the chance level marked. \textbf{b}, Verbalized against internal four-way accuracy per model for both prompts, with the identity and chance lines. \textbf{c}, Per-model internal-minus-verbalized accuracy gap for zero-shot and few-shot prompting, with bootstrap 95\% CIs and the panel-level gaps marked. \textbf{d}, Spearman correlation between the true grade and, respectively, the estimator's scalar projection and the verbalized grade, per model. \textbf{e}, Per model and prompt, the number of held-out claims graded correctly by the estimator only against the number graded correctly by the verbalized answer only, with the identity line. \textbf{f}, Median internal-minus-verbalized accuracy gap by evidence-grade level and prompt, with whiskers showing the standard deviation across models and the held-out claim count per grade; bars are colored in the green palette to distinguish the per-grade quantities from the condition-level legend. \textbf{g}, Per-model zero-shot gap across evidence grades with the cross-model median (dark red line and markers). Estimator quantities are shown in blue and verbalized quantities in red-orange throughout; gaps in panels \textbf{f} and \textbf{g} are in percentage points (pp).}
\label{fig:gap}
\end{figure*}

The internal signal that separates weak from strong evidence is almost entirely absent from what the models say. Asked to grade the evidence behind each claim, a model reaches a four-way accuracy of 22.2 on median zero-shot (range 11.7 to 38.8) and 21.8 with few-shot prompting (range 9.1 to 33.8), at or below the 25.0 expected by chance, against a median of 47.1 for the linear estimator read from the same activations (Fig.~\ref{fig:gap}a). The internal-minus-verbalized accuracy gap is $25.4 \pm 1.7$ [22.0, 28.5] zero-shot and $27.4 \pm 1.7$ [24.1, 30.9] few-shot (paired bootstrap, $p_{\mathrm{FDR}}<0.001$; Fig.~\ref{fig:gap}b,c). The advantage is paired and one-sided at the level of individual claims: a McNemar test is significant after correction in all 31 model-by-prompt comparisons ($p_{\mathrm{FDR}}<10^{-11}$), and in every one the claims graded correctly by the estimator alone outnumber those graded correctly by the verbalized answer alone (median 1{,}126 vs 354; Fig.~\ref{fig:gap}e). Six of the 22 models could not return a usable zero-shot grade at all, collapsing onto a single label or failing to follow the format; they were excluded, so the gap is measured only on the models that did produce a gradeable response and is, if anything, conservative. Per-model verbalized accuracies and internal-minus-verbalized gaps are listed in Table~\ref{tab:behavioral}.

The grade a model states also carries essentially no information about how strong the evidence is. The estimator's scalar projection rises monotonically with the true grade (Spearman correlation, median 0.342, range 0.274 to 0.405, significant in all 20 models), whereas the verbalized grade is uncorrelated with it, at $-0.034$ on median zero-shot (range $-0.088$ to 0.094) and 0.075 few-shot (range $-0.012$ to 0.177) (Fig.~\ref{fig:gap}d). A few of these reach significance given the size of the test set, but their magnitude is an order of magnitude below the estimator's, so a model's explicit evidence grade is effectively flat with respect to the truth even though an ordered grade signal is present in its activations.

The failure is concentrated where it matters most. Broken down by grade, the gap is large and positive at the two weak-evidence levels under both prompting strategies, reaching 40.5 and 39.6 zero-shot and 32.6 and 33.4 few-shot at low and very low evidence, and it is positive there across nearly all of the models (Fig.~\ref{fig:gap}f,g); the moderate and high cells swing in sign between the two prompts (zero-shot $-14.0$ and $50.6$, few-shot $25.0$ and $-19.7$, with only 77 held-out claims in the high cell), driven by the models' tendency to default to particular labels. The models thus separate weakly supported from well supported claims internally, including at the low-evidence end a reader would most need flagged, yet do not report that distinction when asked.

A reader relying on these models would therefore receive weakly supported and well supported claims described with evidence grades that do not track their actual strength, while the information needed to separate them sits recoverable in the model's internal state. What that internal signal is made of, and whether it amounts to more than the surface wording of a claim, is the question we turn to next.

%%%%%%%%%%%%
%%%%%%%%%%%%

%%%%%%%%%%%%%%%%%%%%%%%%%%%%%%%%%%%%%%%%%%%%%%%%%%%%%%%
%%%%%%%%%%%%%%%%%%%%%%%%%%%%%%%%%%%%%%%%%%%%%%%%%%%%%%%
%%%%%%%%%%%%%%%%%%%%%%%%%%%%%%%%%%%%%%%%%%%%%%%%%%%%%%%
%%%%%%%%%%%%%%%%%%%%%%%%%%%%%%%%%%%%%%%%%%%%%%%%%%%%%%%

\subsection*{The recoverable evidence signal is largely lexical}

\begin{figure*}[p]
\centering
\includegraphics[width=0.95\textwidth]{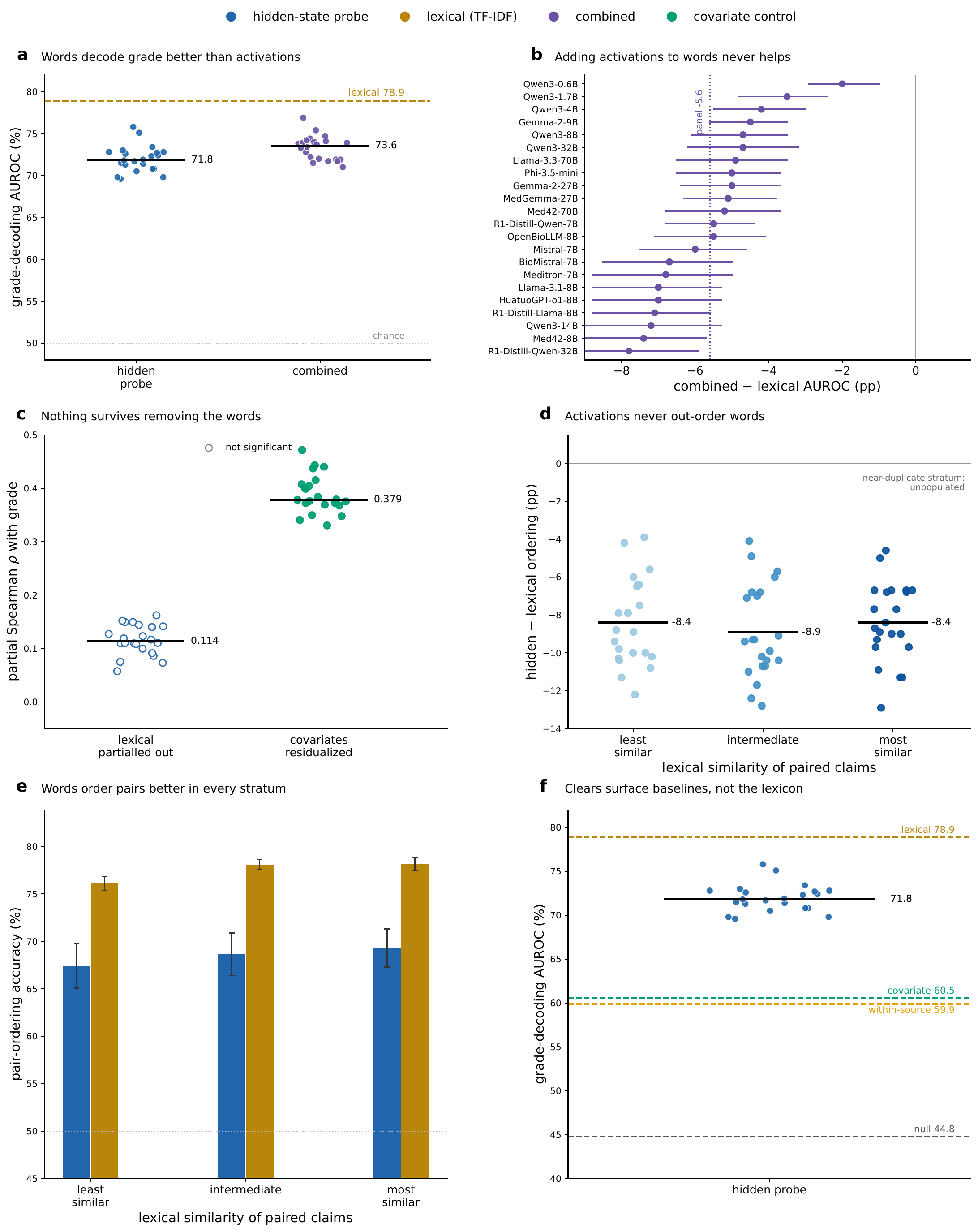}
\caption{Recoverable evidence-grade signal characterized as largely lexical, on the graded held-out test set ($n=3{,}093$ claims) across the 22-model panel. \textbf{a}, Grade-decoding AUROC per model for the hidden-state estimator and the combined estimator, with the model-independent TF-IDF lexical classifier shown as a reference line. \textbf{b}, Per-model combined-minus-lexical AUROC increment, sorted, with bootstrap 95\% CIs and a zero reference. \textbf{c}, Per-model partial Spearman correlation between the hidden grade-score and the true grade after the lexical grade-score is partialled out and, separately, after the length, hedge-density and token-frequency covariates are residualized out. \textbf{d}, Per-model advantage of the hidden grade-score over the lexical score in ordering cross-grade claim pairs, by stratum of lexical similarity between the paired claims, with the near-duplicate stratum noted as unpopulated. \textbf{e}, Pair-ordering accuracy of the hidden and lexical grade-scores in each lexical-similarity stratum. \textbf{f}, Per-model hidden-state estimator grade-decoding AUROC against shuffled-label null, the covariate baseline, the within-source control, and the lexical classifier, shown as reference lines. }
\label{fig:lexical}
\end{figure*}

The evidence-grade signal recovered in the previous sections is one that simple word statistics capture better than the models' internal states. A TF-IDF bag-of-words classifier decodes grade at 78.9, above the per-model decodability of the hidden-state estimator (Table~\ref{tab:decodability}), and combining the two adds nothing: the combined estimator falls below the lexical classifier (difference $-5.6 \pm 0.3$ [$-6.2$, $-5.0$]), and no model gains from the hidden features (0 of 22; Fig.~\ref{fig:lexical}a,b). Reading the hidden states adds nothing to what the words already supply.

To ask whether the representation encodes grade beyond the lexicon at all, we compared pairs of claims carrying different gold grades and tested whether the hidden grade-score orders each pair better than the lexical score. It does not, in any stratum of lexical similarity: the hidden-minus-lexical ordering advantage is $-8.4 \pm 0.5$ [$-9.3$, $-7.4$] for the least similar pairs, $-8.9 \pm 0.5$ [$-9.9$, $-7.9$] at intermediate similarity, and $-8.4 \pm 0.4$ [$-9.2$, $-7.6$] for the most similar pairs, negative and significant in all 22 models in every stratum (paired bootstrap; Fig.~\ref{fig:lexical}d,e). The strictest near-duplicate stratum could not be populated, because the corpus contains almost no near-identical claims with different grades, which is itself part of the finding. Consistent with this, once the lexical grade-score is partialled out the hidden score retains a rank correlation of only 0.114 with the true grade, significant in none of the 22 models (Fig.~\ref{fig:lexical}c).

This is not to say the estimator detects nothing. The hidden signal clears non-lexical surface baselines: it exceeds a classifier built from claim length, hedge density, and token frequency by $11.3 \pm 0.3$ [10.7, 11.9], stays correlated with grade after those covariates are residualized out (partial Spearman 0.379, significant in all 22 models), and within a single grading source, where source identity is held fixed, still decodes grade above chance (median 59.9; Fig.~\ref{fig:lexical}c,f). The signal is therefore not merely hedging, rarity, or length, and not a pure source-identity artifact. What it is not is a representation of evidence strength that extends beyond the words a claim is written in.

The evidence-grade signal these models carry is thus real and robustly recoverable, but largely lexical: it reflects the vocabulary in which a claim is phrased more than any deeper encoding of how well the claim is supported, and almost none of it survives once that vocabulary is accounted for. The behavioral gap of the previous section does not depend on this signal being deep, only on its being recoverable and unspoken. It does raise the question of how far even a lexical signal carries, which we examine next across topics and grading frameworks.

%%%%%%%%%%%%
%%%%%%%%%%%%

\subsection*{The recovered signal does not transfer across topics or grading frameworks}

\begin{figure*}[p]
\centering
\includegraphics[width=\textwidth]{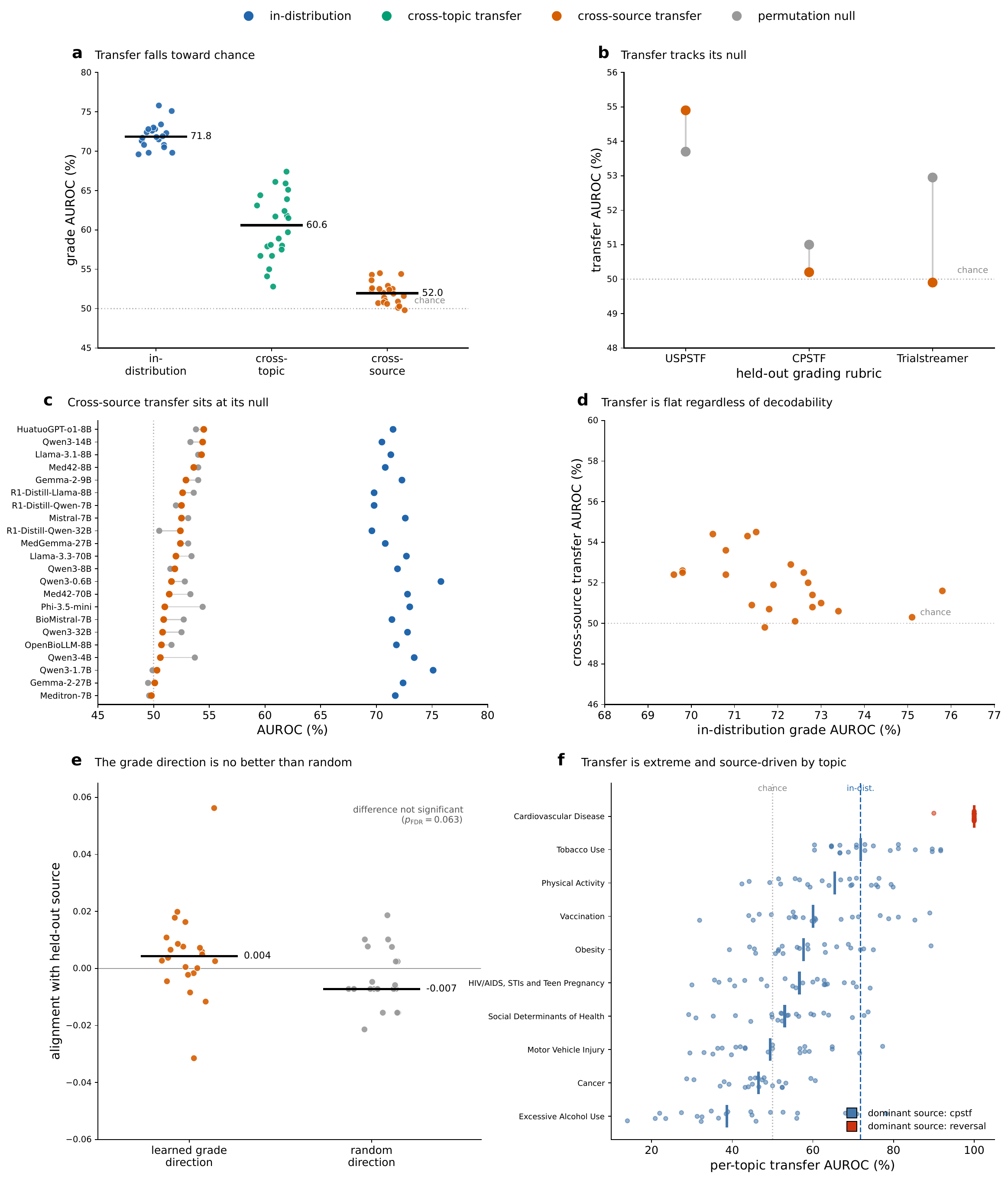}
\caption{Generalization of evidence-grade decoding across grading frameworks and clinical topics, on the graded held-out test set ($n=3{,}093$ claims) across the 22-model panel. \textbf{a}, Per-model in-distribution grade AUROC alongside the cross-topic and cross-grading-source transfer AUROC. \textbf{b}, Leave-one-grading-source-out transfer AUROC against its per-source permutation null, for each held-out rubric. \textbf{c}, Per-model cross-grading-source transfer AUROC against its permutation null and the in-distribution ceiling, sorted. \textbf{d}, Cross-grading-source transfer against in-distribution grade decodability per model, with chance and ceiling references. \textbf{e}, Per-model alignment of the learned grade direction with the held-out grading source and with a random direction. \textbf{f}, Per-topic transfer AUROC across the held-out clinical topics, ordered, colored by the grading source that dominates each topic, with chance and in-distribution references.}
\label{fig:generalize}
\end{figure*}

Consistent with a lexical signal, the estimator does not carry a notion of evidence strength that survives a change of grading framework. Trained on two of the three grading rubrics and tested on the third, it transfers at 52.0 (range 49.8 to 54.5), at chance and, if anything, below the per-source permutation null of 53.1 (transfer-minus-null $-0.6 \pm 0.3$ [$-1.1$, $-0.1$], $p_{\mathrm{FDR}}=0.031$); the in-distribution ceiling of 71.8 sits 20.0 [19.0, 21.1] above the transferred accuracy (Fig.~\ref{fig:generalize}a,c,d). Held out one rubric at a time, the transferred accuracy hovers around its permutation null (Fig.~\ref{fig:generalize}b), and the learned grade direction is no better aligned with the held-out source than a random direction is ($0.009$, $p_{\mathrm{FDR}}=0.063$; Fig.~\ref{fig:generalize}e).

Across topics, the picture is regime-dependent. Holding out whole clinical topics, transfer falls from 71.8 in distribution to a median of 58.0 across models (range 42.6 to 72.4), but that average hides extreme topic dependence: per-topic transfer ranges from 38.7 for excessive alcohol use, below chance, to 100.0 for cardiovascular disease (Fig.~\ref{fig:generalize}f). The high values arise where a held-out topic coincides with a single source and grade structure, so that the estimator is in effect detecting that source instead of grading evidence. The estimator, therefore, generalizes to some topics and not others, in a pattern set by which source a topic's claims happen to come from.

An attempt to recover transfer by projecting the source-identifying subspace out of the representation before refitting did not yield a stable cross-source estimate and is reported as a negative. Taken together, the generalization battery indicates that what the estimator reads is largely the source- and topic-specific vocabulary of a claim, the same lexical signal of the previous section, and not a portable representation of how strong the underlying evidence is.

A signal that does not transfer across the very frameworks used to define evidence strength cannot be read as a general encoding of that property. It also leaves open a related question the corpus is poorly suited to answer cleanly: whether, inside the model, evidence strength is even separable from the plain truth or falsity of a claim. We turn to that next, on data built to make the two independent.

%%%%%%%%%%%%
%%%%%%%%%%%%

\subsection*{Evidence grade is separable from factual truth in the representation}

\begin{figure*}[p]
\centering
\includegraphics[width=\textwidth]{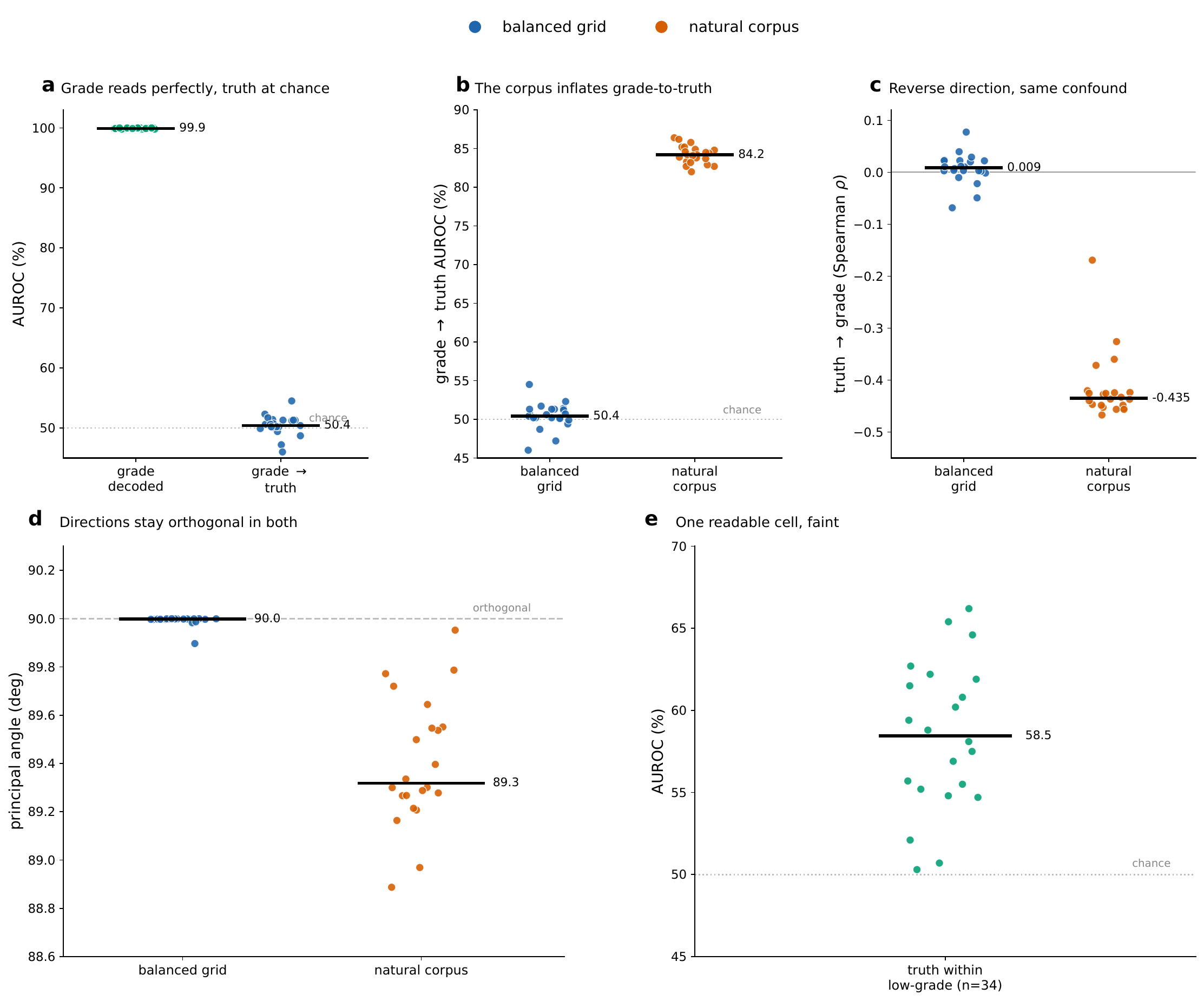}
\caption{Separability of evidence grade and factual truth in the representation, on a balanced grade-by-truth grid that decorrelates the two and, for contrast, on the natural corpus, across the 22-model panel. \textbf{a}, On the balanced grid, grade-decoding area under the receiver operating characteristic curve (AUROC) against the AUROC for predicting truth from the grade direction, with chance marked. \textbf{b}, AUROC for predicting truth from the grade direction, on the balanced grid against the natural corpus, per model. \textbf{c}, Spearman correlation between the truth direction's projection and grade, on the balanced grid against the natural corpus, per model. \textbf{d}, Principal angle between the grade and truth estimator directions, on the balanced grid against the natural corpus, per model. \textbf{e}, Truth-decoding AUROC within low-grade claims on the balanced grid, with chance marked; this is the only decorrelated off-diagonal cell the corpus populates with enough claims to read. Balanced-grid quantities are shown in blue and natural-corpus quantities in orange.}
\label{fig:dissociation}
\end{figure*}

On data constructed to make grade and truth statistically independent, the two are distinct, near-orthogonal directions in the representation, and neither predicts the other. With grade and truth decorrelated by design (Cramer's V 0.000 by construction), the grade and truth estimator directions are orthogonal (cosine 0.000, principal angle $90.0^\circ$; Fig.~\ref{fig:dissociation}d), the grade direction predicts truth at chance ($50.4 \pm 0.4$ [49.7, 51.1], above chance in 0 of 22 models; Fig.~\ref{fig:dissociation}a), and the truth direction carries essentially no rank information about grade ($0.009$, significant in 0 of 22 models; Fig.~\ref{fig:dissociation}c). Grade itself is decoded at near-ceiling on this grid (99.9; Fig.~\ref{fig:dissociation}a), so the absent cross-prediction reflects genuine separation. Evidence grade is, in other words, not a relabeling of whether a claim is true.

The one place the two meet is faint and local. Within low-grade claims alone, a truth estimator reaches a median of 58.5 (range 50.3 to 66.2), above chance in 18 of 22 models, though on small samples (34 claims; Fig.~\ref{fig:dissociation}e). This is the only off-diagonal cell the corpus populates with enough decorrelated claims to read: the high-grade truth cell and the grade-within-truth cells are fully source-confounded, separable at near-100 for trivial reasons, and are excluded from the analysis. We therefore restrict any positive grade-truth link to the weakest-evidence claims and do not claim a general one.

The natural corpus, by contrast, makes grade and truth look tightly coupled, which is exactly the trap the balanced design avoids. There the two labels are strongly associated (Cramer's V 0.614, $p_{\mathrm{FDR}}=9.3\times10^{-51}$; Fig.~\ref{fig:corpus}e), and the grade direction appears to predict truth well ($84.2 \pm 0.2$ [83.7, 84.7], above chance in all 22 models; Fig.~\ref{fig:dissociation}b). But this is the association speaking, not a shared code: the grade and truth directions are just as orthogonal in the corpus (cosine 0.012, principal angle $89.3^\circ$; Fig.~\ref{fig:dissociation}d) as on the balanced grid, and that same grade direction drops to chance (50.4) once the two are decorrelated (Fig.~\ref{fig:dissociation}b). The apparent identity of grade and truth is a property of how the corpus is distributed, not of the model.

Evidence grade is thus a distinct internal signal from factual correctness, not a correctness detector under another name, even though the corpus's strong grade-truth coupling would suggest otherwise and only the low-grade cell can be read without confounding. What remains untested is whether this recoverable grade signal can be made to change a model's behavior at all, which we examine next with activation steering.

%%%%%%%%%%%%
%%%%%%%%%%%%

%%%%%%%%%%%%%%%%%%%%%%%%%%%%%%%%%%%%%%%%%%
%%%%%%%%%%%%%%%%%%%%%%%%%%%%%%%%%%%%%%%%%%

\subsection*{Steering along the grade direction does not change what the models express}

The recoverable grade direction is not a usable lever on what a model expresses. Steering generation along the grade direction does not move a model's stated certainty more than a matched random direction does: across the six steered models, the grade-minus-random coherent slope is $-0.456$ ($p_{\mathrm{FDR}}=0.393$), with the grade slope exceeding the random slope in only three of the six models (Supplementary Fig.~\ref{sfig:steering}). Some models' outputs shift under steering, but the grade direction confers no specific control over the certainty they express, and we make no causal claim from it.

The grade signal these models carry is therefore recoverable from their activations but neither stated when the models are asked nor reliably moved when the activations are pushed along it. The operative finding remains the behavioral gap: the information a reader would need to weight a clinical claim is present inside the model and absent from its output.

%%%%%%%%%%%%%%%%%%%%%%%%%%%%%%%%%%%%%%%%%%
%%%%%%%%%%%%%%%%%%%%%%%%%%%%%%%%%%%%%%%%%%

\subsection*{The internal signal can be extracted to flag weakly supported claims}

\begin{figure*}[p]
\centering
\includegraphics[width=\textwidth]{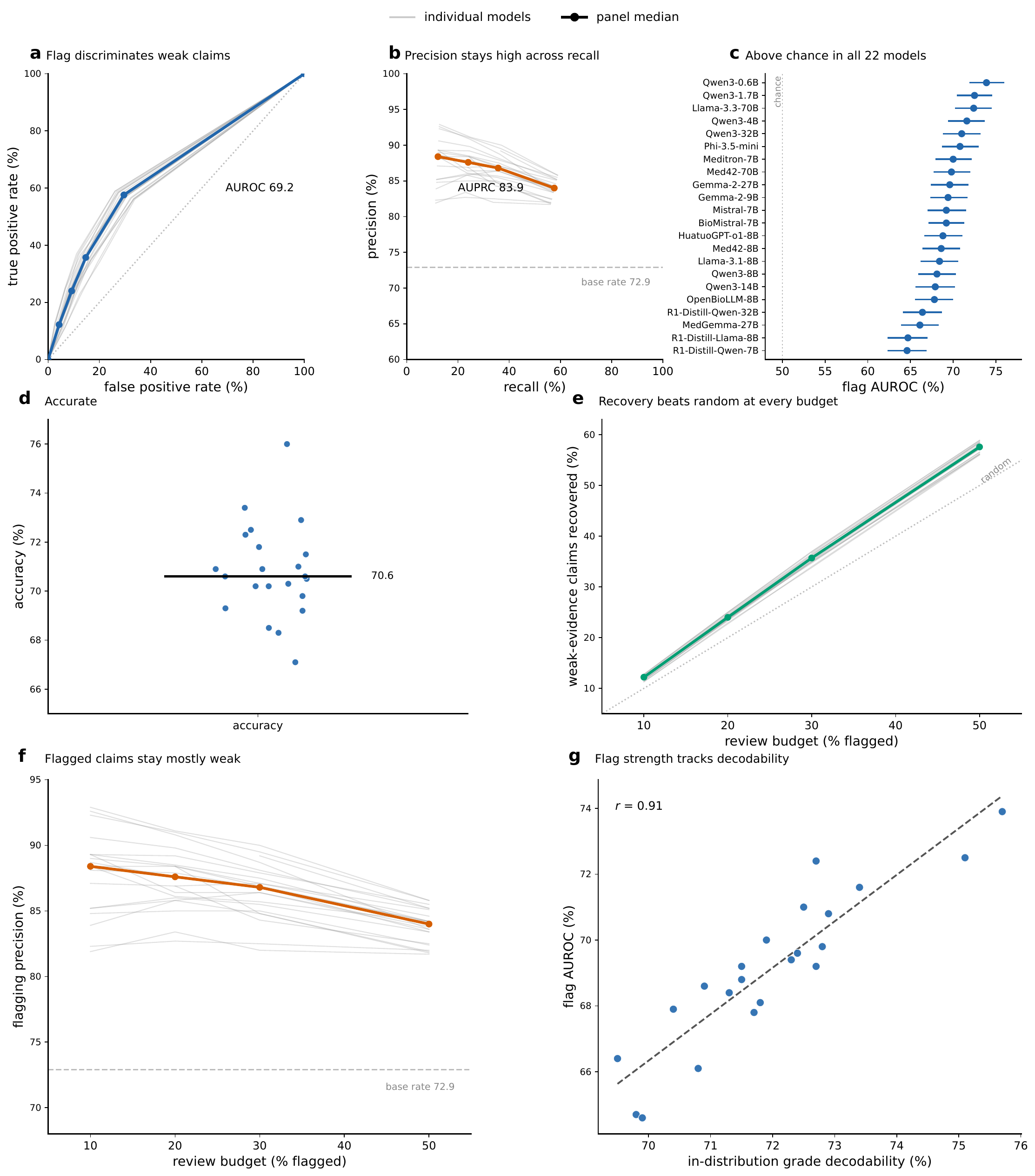}
\caption{An external estimator flags weakly supported (low and very-low grade) claims from clinical-LLM hidden states, on the graded held-out test set ($n=3{,}093$ claims) across the 22-model panel. \textbf{a}, Operating points of the weak-evidence flag in receiver-operating-characteristic coordinates, at flag rates of 10, 20, 30 and 50\%, per model and at the panel median, with the chance diagonal. \textbf{b}, The same operating points in precision-recall coordinates, with the weak-evidence base rate shown as the no-skill line. \textbf{c}, Per-model flag AUROC, sorted, with bootstrap 95\% CIs, against chance. \textbf{d}, Per-model flagging accuracy and Brier score, on separate axes. \textbf{e}, The fraction of weak-evidence claims recovered as a function of the review budget, the rate at which claims are flagged, against the random-flagging diagonal. \textbf{f}, Flagging precision as a function of the review budget, against the base rate. \textbf{g}, Flag AUROC against in-distribution grade decodability per model.}
\label{fig:triage}
\end{figure*}

The signal the models will not verbalize can be recovered by an external estimator and put to use. An estimator trained to flag the weak-evidence claims (low and very low grade) reaches a median AUROC of 69.2 (range 64.6 to 73.9) and an AUPRC of 83.9 (80.3 to 86.9), above chance in all 22 models, at a median accuracy of 70.6 (67.1 to 76.0) and a Brier score between 0.2 and 0.3 (Fig.~\ref{fig:triage}a--d; per-model flag AUROC in Table~\ref{tab:behavioral}). Across review budgets it recovers a substantial share of the weak-evidence claims at a precision that stays well above the base rate (Fig.~\ref{fig:triage}e,f), and a model's flagging strength tracks its decodability (Fig.~\ref{fig:triage}g). Like the signal it rests on, this flag is largely lexical, so a plain text classifier reaches similar accuracy; the point is that the capability is cheaply available from the model even though the model does not surface it.

The information needed to flag a weakly supported clinical claim is therefore recoverable from a model's activations, or from its text, at useful accuracy, yet absent from the graded answers the model gives when asked. Using these models safely means reading the signal out, not trusting the grades they state.

%%%%%%%%%%%%
%%%%%%%%%%%%

\begin{table}[t]
\centering
\caption{Per-model behavioral and characterization results, on the graded held-out test set ($n=3{,}093$ claims). Models are grouped by family as in Table~\ref{tab:decodability}. Verbalized accuracy is the model's zero-shot four-way grade accuracy and Gap is the internal-estimator-minus-verbalized accuracy difference, both for the 16 models that produced a parseable zero-shot grade (N/A otherwise). Combined$-$lexical is the change in grade AUROC from adding the hidden features to the lexical classifier. Source transfer is the leave-one-grading-source-out transfer AUROC, and Flag AUROC is the area under the curve for flagging weak-evidence claims. All values are percentages, reported as the mean with its 95\% CI in brackets over $10{,}000$ bootstrap resamples. N/A, not applicable.}
\label{tab:behavioral}
\scriptsize
\setlength{\tabcolsep}{6pt}
\renewcommand{\arraystretch}{1.15}
\begin{tabular}{@{}lccccc@{}}
\toprule
Model & Verbalized acc. & Gap & Combined$-$lexical & Source transfer & Flag AUROC \\
 & (zero-shot) & (internal $-$ verb.) & ($\Delta$ AUROC) & (AUROC) & (AUROC) \\
\midrule
\multicolumn{6}{@{}l}{\textit{General-domain}} \\
Qwen3-0.6B & N/A & N/A & $-2.0$~[$-2.9$,~$-1.0$] & 51.6 [50.6, 52.8] & 73.9 [72.0, 75.9] \\
Qwen3-1.7B & N/A & N/A & $-3.5$~[$-4.8$,~$-2.4$] & 50.3 [44.7, 55.0] & 72.5 [70.5, 74.5] \\
Qwen3-4B & N/A & N/A & $-4.2$~[$-5.5$,~$-3.0$] & 50.6 [49.3, 52.5] & 71.6 [69.5, 73.6] \\
Qwen3-8B & 11.9 [10.5, 13.3] & 33.8 [31.4, 36.4] & $-4.7$~[$-6.1$,~$-3.5$] & 51.9 [46.9, 59.4] & 68.1 [66.0, 70.2] \\
Qwen3-14B & N/A & N/A & $-7.2$~[$-9.0$,~$-5.3$] & 54.4 [50.3, 60.3] & 67.9 [65.7, 70.1] \\
Qwen3-32B & 24.1 [22.5, 25.6] & 23.2 [20.8, 25.6] & $-4.7$~[$-6.2$,~$-3.2$] & 50.8 [47.1, 56.5] & 71.0 [68.9, 73.1] \\
Phi-3.5-mini & 28.1 [26.5, 29.8] & 19.8 [17.4, 22.2] & $-5.0$~[$-6.5$,~$-3.7$] & 51.0 [48.1, 54.6] & 70.8 [68.8, 72.9] \\
Mistral-7B-v0.3 & 23.2 [21.7, 24.7] & 24.1 [21.8, 26.4] & $-6.0$~[$-7.5$,~$-4.6$] & 52.5 [49.7, 57.5] & 69.2 [67.1, 71.4] \\
Llama-3.1-8B & 23.6 [22.1, 25.1] & 23.6 [21.2, 25.9] & $-7.0$~[$-8.8$,~$-5.3$] & 54.3 [50.4, 56.7] & 68.4 [66.3, 70.5] \\
Gemma-2-9B & 23.1 [21.6, 24.6] & 23.4 [21.0, 25.7] & $-4.5$~[$-5.6$,~$-3.5$] & 52.9 [48.6, 58.6] & 69.4 [67.4, 71.6] \\
Gemma-2-27B & 21.4 [19.9, 22.8] & 25.7 [23.4, 28.0] & $-5.0$~[$-6.4$,~$-3.7$] & 50.1 [46.5, 53.6] & 69.6 [67.5, 71.7] \\
Llama-3.3-70B & 17.3 [15.9, 18.6] & 31.8 [29.6, 34.1] & $-4.9$~[$-6.5$,~$-3.5$] & 52.0 [48.1, 54.8] & 72.4 [70.3, 74.4] \\
\midrule
\multicolumn{6}{@{}l}{\textit{Medical domain-adapted}} \\
BioMistral-7B & 20.3 [18.9, 21.8] & 25.5 [23.2, 27.8] & $-6.7$~[$-8.5$,~$-5.0$] & 50.9 [49.3, 53.3] & 69.2 [67.2, 71.2] \\
Meditron-7B & 38.8 [37.1, 40.6] & 9.0 [6.5, 11.4] & $-6.8$~[$-8.8$,~$-5.0$] & 49.8 [46.6, 51.6] & 70.0 [68.0, 72.1] \\
OpenBioLLM-8B & 29.1 [27.5, 30.7] & 17.3 [14.9, 19.6] & $-5.5$~[$-7.1$,~$-4.1$] & 50.7 [48.0, 54.3] & 67.8 [65.6, 69.9] \\
Med42-8B & 24.7 [23.2, 26.2] & 21.5 [19.1, 23.9] & $-7.4$~[$-9.2$,~$-5.7$] & 53.6 [50.9, 55.9] & 68.6 [66.5, 70.7] \\
MedGemma-27B & 14.8 [13.5, 16.1] & 30.6 [28.5, 32.7] & $-5.1$~[$-6.3$,~$-3.8$] & 52.4 [48.3, 57.1] & 66.1 [64.0, 68.2] \\
Med42-70B & 11.7 [10.6, 12.9] & 35.1 [33.0, 37.1] & $-5.2$~[$-6.8$,~$-3.7$] & 51.4 [48.5, 53.6] & 69.8 [67.8, 71.9] \\
\midrule
\multicolumn{6}{@{}l}{\textit{Reasoning-distilled}} \\
R1-Distill-Qwen-7B & 12.5 [11.0, 13.9] & 31.8 [29.3, 34.4] & $-5.5$~[$-6.8$,~$-4.4$] & 52.5 [47.7, 59.7] & 64.6 [62.4, 66.8] \\
R1-Distill-Llama-8B & N/A & N/A & $-7.1$~[$-8.8$,~$-5.6$] & 52.6 [49.9, 54.2] & 64.7 [62.4, 66.9] \\
R1-Distill-Qwen-32B & N/A & N/A & $-7.8$~[$-9.8$,~$-5.9$] & 52.4 [49.2, 58.1] & 66.4 [64.2, 68.6] \\
HuatuoGPT-o1-8B & 17.0 [15.7, 18.3] & 30.2 [28.0, 32.5] & $-7.0$~[$-8.8$,~$-5.3$] & 54.5 [49.8, 56.9] & 68.8 [66.7, 71.0] \\
\bottomrule
\end{tabular}
\end{table}

\section*{Discussion}

We asked whether frozen, open-weight clinical language models internally register how strongly a claim is supported by evidence, and whether they communicate that strength when prompted. The answer divides the two cleanly. An ordered evidence-grade signal is linearly recoverable from the hidden states of every model we examined, yet the grade a model states when asked is uninformative about how strong the underlying evidence is, so the information a reader would need to weigh a clinical assertion sits inside the model and is missing from what it says. As these systems move from benchmark demonstrations into clinical documentation, evidence summarization, and patient-facing answers \cite{singhal2023large,thirunavukarasu2023large,vanveen2024adapted,singhal2025expert}, it is this dissociation, and not any single decoding accuracy, that bears on whether their output can be used safely.

The finding sharpens a problem that is already recognized. Language models communicate certainty poorly: their stated confidence is weakly calibrated and frequently fails to track whether an answer is correct \cite{kadavath2022language,lin2022teaching,tian2023just,xiong2024can}, and the explanations they offer can leave a reader more assured than the model's own internal estimate warrants \cite{steyvers2025calibration}. Our result locates the failure more precisely than miscalibration alone would. The shortfall is not that the output signal is merely noisy; an ordered representation of evidentiary strength is linearly present in the activations and effectively absent from the verbalized grade. This carries the observation that internal states encode more about a model's reliability than its outputs reveal \cite{orgad2025llms} from factual truth to graded strength of support. A practical corollary follows: confidence-elicitation methods that operate on the output, whether verbalized probabilities, agreement across samples \cite{farquhar2024detecting}, or features of an explanation, cannot recover a signal the output does not contain, and only the representation can. In this setting, calibration is partly a problem of expression, and not solely one of knowledge.

What the estimator recovers is nonetheless shallower than the language of internal `knowledge' tends to suggest. A substantial line of interpretability research treats the linear decodability of a property as evidence that a model represents it \cite{azaria2023internal,burns2023discovering,li2023inference,marks2024geometry}. Our controls qualify what decodability licenses here. Word statistics alone decode evidence grade better than the hidden states do, almost none of the hidden signal survives once the lexical grade-score is partialled out, and the estimator does not transfer to held-out clinical topics or to grading frameworks it was not trained on. The model has thus not acquired a portable, abstract representation of evidence strength; the estimator reads, in the main, the vocabulary a claim is phrased in. This accords with recent findings that estimators can capture surface lexical regularities and memorized statistical associations, and that such signals fail to generalize across datasets \cite{cheang2026recall,orgad2025llms}. The methodological lesson reaches past this study: a decodability result reported without lexical and out-of-distribution controls overstates what is encoded \cite{hewitt2019designing}, and the claim that a model represents a construct should be reserved for a signal that survives them.

How decodability varied across the panel was itself unexpected. It did not increase with model capacity; it was strongest in the smallest model and weakest in the larger and the reasoning-distilled ones, while medical domain adaptation left it largely unchanged. The lexical nature of the signal suggests one reading. A small model remains close to surface token statistics that a linear estimator can read directly, whereas added scale, instruction tuning, and reasoning distillation tend to reshape representations toward more distributed and task-aligned features that a linear readout at the final token reaches less well \cite{deepseekai2025r1,chen2024huatuogpto1}. This is a statement about linear accessibility and not about what a model can compute: a larger model may use the same surface cues without exposing them to an estimator. The implication runs counter to intuition. A representation-level safety tool may perform best on the smallest models, and choosing a clinical model for the apparent cleanliness of its internal evidence signal is not warranted \cite{wind2026safetyaccuracyfollowdifferent}.

Evidence strength further proved to be an internal property distinct from factual correctness, the property on which most representational work on model `knowledge' has concentrated \cite{azaria2023internal,burns2023discovering,marks2024geometry,li2023inference,chen2024inside}. On data that decorrelates the two by construction, the grade direction predicts truth no better than chance and lies nearly orthogonal to the truth direction, even though grade and truth appear strongly coupled in the natural corpus. The distinction is clinically real. Well-supported recommendations are sometimes overturned, and catalogued reversals of established practice make the gap between well-supported and true concrete \cite{herrera2019reversals}: a correctness detector would not flag a strongly supported recommendation that later reverses, whereas a signal of evidentiary strength speaks to how much weight a claim should carry on present knowledge \cite{guyatt2008grade,uspstf2018procedure}. The apparent identity of grade and truth in the unbalanced corpus is a property of how clinical evidence is distributed, and it illustrates how an association read from natural data can pass for a shared internal code when it is not.

The dissociation still admits a constructive use. Because the grade signal is recoverable, a single forward pass can be converted into an external flag for weakly supported claims that needs no retrieval and no outside evidence \cite{wind2025multi,tayebi2025radiorag}, at an accuracy adequate to decide which of a model's assertions a clinician should scrutinize \cite{hager2024evaluation,vanveen2024adapted}. That a plain text classifier reaches comparable accuracy does not undercut the case; the capability is cheaply available beside the model, even though the model will not surface it. Acting on the representation directly was what failed: although directions in activation space can sometimes steer behavior \cite{zou2023representation,turner2023activation}, the grade direction proved readable but not a lever on what the model stated, leaving the internal signal dissociated from behavior even under intervention; this is consistent with evidence that activation steering is frequently unreliable and input-dependent \cite{tan2024analysing}, and we draw no causal conclusion from it. The operational reading is therefore direct. The grades these models state should not be taken as measures of evidentiary strength, and an inexpensive downstream check, an estimator, or an equally cheap classifier over the claim text, belongs between a model's confident phrasing and the clinician who relies on it \cite{singhal2025expert}.

Several limitations bound these conclusions. First, the recovered signal is largely lexical, so what the estimator reads is closer to a surface correlate of evidence strength than to a deep representation of it \cite{geirhos2020shortcut,gururangan2018annotation,lotfinia2026visionlanguagemodelschestradiography}; corpora that decouple grade from wording, through paraphrase control or adversarially matched claims, would test more stringently for any non-lexical encoding \cite{arasteh2026casegroundedevidenceverificationframework}. Second, three native grading systems and the reversal grades were harmonized onto a single four-level scale, a mapping that involves judgment and could shift absolute decodability, and although we report a sensitivity analysis on the Trialstreamer rigor bands, a fully framework-native treatment would be stronger. Third, the behavioral gap was measured against particular zero-shot and few-shot prompts \cite{sclar2024quantifying,tayebi2026framing} and several models returned no parseable grade, so the gap is an upper bound on what these prompts surface and not proof that the grade cannot be expressed; prompts designed to elicit grades, or light fine-tuning to report them, might narrow it. Fourth, the separability of grade from truth rests on a balanced grid whose only confound-free off-diagonal cell is small and limited to the weakest-evidence claims, with the remaining cells confounded by source, so corpora deliberately enriched for strong-but-false and weak-but-true claims would be needed to read the dissociation across all grades. Fifth, we studied open-weight models with frozen weights and a final-token linear estimator, and other architectures, layers, readout positions, or nonlinear estimators could expose the signal differently \cite{belinkov2022probing}, while closed models reachable only through an interface remain untested. Sixth, the endpoints throughout are decoding and elicitation metrics on atomic claims instead of clinical decisions or clinician trust, so whether reading the signal out improves real review is a question for prospective evaluation with clinicians before any deployment.

Clinical language models carry an ordered signal of how strongly a claim is supported, one recoverable from their activations even though it is in large part a trace of the words a claim is written in, and they do not state that strength when asked. The fact that matters for safety is the distance between what can be recovered and what is said. Until such models are built or guided to express evidentiary strength faithfully, the strength of a clinical claim should be read from the representation or from the text by an external check, not accepted from the grade the model itself reports. Making evidence strength something a model both encodes more than superficially and conveys faithfully is a concrete target for the next generation of clinical systems, and a precondition for letting their answers carry the evidentiary weight that clinical judgment demands.

%%%%%%%%%%%%%%%%%%%%%%%%%%%%%%%%%%%%%%%%%%%%%%%%%%%%%%%
%%%%%%%%%%%%%%%%%%%%%%%%%%%%%%%%%%%%%%%%%%%%%%%%%%%%%%%
%%%%%%%%%%%%%%%%%%%%%%%%%%%%%%%%%%%%%%%%%%%%%%%%%%%%%%%
%%%%%%%%%%%%%%%%%%%%%%%%%%%%%%%%%%%%%%%%%%%%%%%%%%%%%%%

\section*{Data availability}

All evidence-grade data used in this study come from publicly available sources, and no data were redistributed here. The USPSTF recommendations were downloaded directly from the public USPSTF website, namely the searchable index of all published recommendations at \url{https://www.uspreventiveservicestaskforce.org/uspstf/topic_search_results?topic_status=P} together with the linked per-topic recommendation pages, under its license. The CPSTF Community Guide findings were obtained by direct download of the official all-active-findings listing at \url{https://www.thecommunityguide.org/media/pdf/cpstf-finding-lists/CPSTF-All-Findings-508.pdf} (public). The medical reversals were taken from Herrera-Perez et al. \cite{herrera2019reversals}.
The Trialstreamer database \cite{marshall2020trialstreamer} is available as a bulk download archived on Zenodo (concept DOI \url{https://doi.org/10.5281/zenodo.3767068}; the snapshot used here is \url{https://zenodo.org/records/4066403}) under its license. The EBM-NLP corpus \cite{nye2018ebmnlp} is available at \url{https://github.com/bepnye/EBM-NLP} under its license, and the Evidence Inference corpus \cite{lehman2019inferring} at \url{https://github.com/jayded/evidence-inference}, both as direct public downloads released for research use. The claim texts from these three trial-literature sources (Trialstreamer, Evidence Inference, and EBM-NLP) derive from copyrighted PubMed abstracts; accordingly we release no derived corpus and instead document the full extraction and harmonization pipeline (Supplementary Note~\ref{snote:datacuration}) so that the corpus can be regenerated from these sources.

%%%%%%%%%%%%%%%%%%%%%%%%%%%%%%%%%%%%%%%%%%%%%%%%%%%%%%%
%%%%%%%%%%%%%%%%%%%%%%%%%%%%%%%%%%%%%%%%%%%%%%%%%%%%%%%
%%%%%%%%%%%%%%%%%%%%%%%%%%%%%%%%%%%%%%%%%%%%%%%%%%%%%%%
%%%%%%%%%%%%%%%%%%%%%%%%%%%%%%%%%%%%%%%%%%%%%%%%%%%%%%%

\section*{Code availability}

The analysis code is publicly available at \url{https://github.com/tayebiarasteh/episteme}; the repository provides the data-build and analysis code and does not redistribute the model weights or the underlying datasets. All 22 models are local, open-weight models run entirely on-site without any cloud or third-party API. Model access differs by checkpoint: the Qwen3, Mistral-7B, Phi-3.5-mini, BioMistral-7B, and DeepSeek-R1-Qwen-distilled checkpoints are ungated downloads under permissive licenses, whereas the Llama-3.1, Llama-3.3, Gemma-2, MedGemma, and the Llama- or Gemma-derived medical and distilled checkpoints (Meditron, OpenBioLLM, Med42, HuatuoGPT-o1, and R1-Distill-Llama) carry custom licenses whose acceptance is required before use, with several gated on the Hugging Face Hub behind an access request. Experiments ran in June 2026. The Hugging Face URLs of the 22 model checkpoints are:

\textit{General-domain models:}
\begin{itemize}
  \item Qwen3-0.6B: \url{https://huggingface.co/Qwen/Qwen3-0.6B}
  \item Qwen3-1.7B: \url{https://huggingface.co/Qwen/Qwen3-1.7B}
  \item Qwen3-4B: \url{https://huggingface.co/Qwen/Qwen3-4B}
  \item Qwen3-8B: \url{https://huggingface.co/Qwen/Qwen3-8B}
  \item Qwen3-14B: \url{https://huggingface.co/Qwen/Qwen3-14B}
  \item Qwen3-32B: \url{https://huggingface.co/Qwen/Qwen3-32B}
  \item Phi-3.5-mini: \url{https://huggingface.co/microsoft/Phi-3.5-mini-instruct}
  \item Mistral-7B-v0.3: \url{https://huggingface.co/mistralai/Mistral-7B-Instruct-v0.3}
  \item Llama-3.1-8B: \url{https://huggingface.co/meta-llama/Llama-3.1-8B-Instruct}
  \item Gemma-2-9B: \url{https://huggingface.co/google/gemma-2-9b-it}
  \item Gemma-2-27B: \url{https://huggingface.co/google/gemma-2-27b-it}
  \item Llama-3.3-70B: \url{https://huggingface.co/meta-llama/Llama-3.3-70B-Instruct}
\end{itemize}
\textit{Medical domain-adapted models:}
\begin{itemize}
  \item BioMistral-7B: \url{https://huggingface.co/BioMistral/BioMistral-7B}
  \item Meditron-7B: \url{https://huggingface.co/epfl-llm/meditron-7b}
  \item OpenBioLLM-8B: \url{https://huggingface.co/aaditya/OpenBioLLM-Llama3-8B}
  \item Med42-8B: \url{https://huggingface.co/m42-health/Llama3-Med42-8B}
  \item MedGemma-27B: \url{https://huggingface.co/google/medgemma-27b-it}
  \item Med42-70B: \url{https://huggingface.co/m42-health/Llama3-Med42-70B}
\end{itemize}
\textit{Reasoning-distilled models:}
\begin{itemize}
  \item R1-Distill-Qwen-7B: \url{https://huggingface.co/deepseek-ai/DeepSeek-R1-Distill-Qwen-7B}
  \item R1-Distill-Llama-8B: \url{https://huggingface.co/deepseek-ai/DeepSeek-R1-Distill-Llama-8B}
  \item R1-Distill-Qwen-32B: \url{https://huggingface.co/deepseek-ai/DeepSeek-R1-Distill-Qwen-32B}
  \item HuatuoGPT-o1-8B: \url{https://huggingface.co/FreedomIntelligence/HuatuoGPT-o1-8B}
\end{itemize}
Analyses used: Python~3.11, NumPy~1.26, SciPy~1.17, pandas~3.0, scikit-learn~1.8, PyTorch~2.9 (CUDA~13.0), Hugging Face Transformers~5.0, tokenizers~0.22, huggingface-hub~1.3, safetensors~0.7, and bitsandbytes~0.49 for 4-bit quantization of the six models of 27 billion parameters and above. Hidden-state extraction ran on multiple NVIDIA L40S GPUs (48~GB VRAM each; Intel Xeon Silver 4310 CPUs); at least two such GPUs are required to run the 70-billion-parameter models in 4-bit quantization.

\section*{Acknowledgements}

STA is supported by the Excellence Strategy of the German Federal Government, the Länder, and RWTH ERS (START\_526-26).

%%%%%%%%%%%%%%%%%%%%%%%%%%%%%%%%%%%%%%%%%%%%%%%%%%%%%%%
%%%%%%%%%%%%%%%%%%%%%%%%%%%%%%%%%%%%%%%%%%%%%%%%%%%%%%%
%%%%%%%%%%%%%%%%%%%%%%%%%%%%%%%%%%%%%%%%%%%%%%%%%%%%%%%
%%%%%%%%%%%%%%%%%%%%%%%%%%%%%%%%%%%%%%%%%%%%%%%%%%%%%%%

\section*{Author contributions}

STA designed and performed the study.

%%%%%%%%%%%%%%%%%%%%%%%%%%%%%%%%%%%%%%%%%%%%%%%%%%%%%%%
%%%%%%%%%%%%%%%%%%%%%%%%%%%%%%%%%%%%%%%%%%%%%%%%%%%%%%%
%%%%%%%%%%%%%%%%%%%%%%%%%%%%%%%%%%%%%%%%%%%%%%%%%%%%%%%
%%%%%%%%%%%%%%%%%%%%%%%%%%%%%%%%%%%%%%%%%%%%%%%%%%%%%%%

\section*{Competing interests}

STA is on the editorial board of Communications Medicine and of European Radiology Experimental, and on the trainee editorial board of Radiology: Artificial Intelligence. The author does not have any other competing interests to disclose.

%%%%%%%%%%%%%%%%%%%%%%%%%%%%%%%%%%%%%%%%%%%%%%%%%%%%%%%
%%%%%%%%%%%%%%%%%%%%%%%%%%%%%%%%%%%%%%%%%%%%%%%%%%%%%%%
%%%%%%%%%%%%%%%%%%%%%%%%%%%%%%%%%%%%%%%%%%%%%%%%%%%%%%%
%%%%%%%%%%%%%%%%%%%%%%%%%%%%%%%%%%%%%%%%%%%%%%%%%%%%%%%

\bibliographystyle{splncs04}
\bibliography{bibliography}

\clearpage

\setcounter{table}{0}
\setcounter{figure}{0}
\setcounter{equation}{0}
\renewcommand{\tablename}{Supplementary Table}
\renewcommand{\figurename}{Supplementary Fig.}
\floatname{algorithm}{Supplementary Algorithm}
\renewcommand{\thealgorithm}{\arabic{algorithm}}
\renewcommand{\theequation}{S\arabic{equation}}

\section*{Supplementary information}

\section*{Supplementary Note 1: Clinical claim corpus construction.}
\label{snote:datacuration}

This note documents the full corpus-construction pipeline end to end, from the six raw public sources to the assembled, split corpus of $45{,}134$ claims read by all experiments. The per-source and cross-tabulated statistics are collected in Supplementary Tables~\ref{stab:sourcepipeline}, \ref{stab:gradesource}, and \ref{stab:sourcetruth}.

\subsection*{Overall pipeline}

Each source is read by a dedicated loader that emits a per-source table in a single canonical schema: a stable claim identifier, the claim text, the source name, the clinical specialty where available, the harmonized four-level grade and the raw native grade string it was mapped from, the factual-truth label, and a crossing-cell tag marking the rare claims for which grade and truth are decoupled. The per-source tables are then concatenated and passed through one shared assembly path: truth labels are coerced to lowercase strings; claim text is whitespace-normalized and empty rows dropped; exact duplicate claim texts are removed across the entire corpus; three surface covariates are computed (claim length in words, a hedge-term density from a fixed hedging lexicon, and a token-frequency proxy); degenerate non-claims are filtered out; and the corpus is partitioned into training, validation, and test splits. Claim identifiers are verified unique and the corpus is written once. The factual-truth assignment rule, applied uniformly across sources, is the one stated in the Methods (true for current best-evidence recommendations, false for practices a trial overturned, truth-uncertain for insufficient or single-trial evidence).

\subsection*{Per-source extraction}

\paragraph{USPSTF (122 claims).} US Preventive Services Task Force recommendations were downloaded from the public USPSTF website by parsing the searchable index of all published recommendations, which lists every recommendation topic together with its letter grade or grades (A, B, C, D, and I) across the full range; no API or credentials were used. The letter grade was harmonized as A~$\rightarrow$~high, B~$\rightarrow$~moderate, C~$\rightarrow$~low, and D or I~$\rightarrow$~very low. Topics that assign different grades to different subpopulations were resolved by reading the topic's recommendation-summary page and emitting one neutral claim per subpopulation and grade, with the claim text built from the topic and service type alone so that the grade word never appears in the claim. USPSTF is the only source that spans the full grade range within a single rubric and writing style, which is what lets it serve as the held-out target of the cross-framework transfer analysis. Its grade distribution is high~12, moderate~35, low~7, and very~low~68 (Supplementary Table~\ref{stab:gradesource}).

\paragraph{CPSTF (224 claims).} Community Preventive Services Task Force findings were parsed from the official ``All Active Findings'' PDF, published by the Community Guide \cite{cpstf2024community}. The PDF is a ruled three-column table (intervention, finding, date) whose intervention names frequently wrap across several lines; native table extraction was used instead of left-to-right word reconstruction, because the latter interleaves the finding column into multi-line names and garbles them. The indentation of each name cell distinguishes topic-band headers, section headers, top-level findings, and indented sub-rows (each composed as ``parent header: sub-row name''), and header lines ending in a continuation word were merged so that a name wrapped across two header lines is kept whole. Findings map as recommended at strong strength~$\rightarrow$~high, recommended at sufficient or unstated strength~$\rightarrow$~moderate, recommended-against at strong strength~$\rightarrow$~high, recommended-against without a strength qualifier~$\rightarrow$~moderate, and insufficient-evidence~$\rightarrow$~low, with each claim phrased as an effectiveness statement (``$\langle$intervention$\rangle$ is an effective community health intervention for $\langle$topic$\rangle$''). A recommended-against finding keeps the effectiveness phrasing but carries a false truth label, which is the intended grade-truth decoupling. The current edition contains only two recommended-against findings, so CPSTF contributes essentially true high- and moderate-grade claims and is valuable as a third independent grading rubric and not a source of false claims. Its grade distribution is high~106, moderate~68, and low~50.

\paragraph{Medical reversals (395 claims).} Reversals of established practice, each a once-standard practice that a later randomized trial overturned, were read from the eLife reversals supplement \cite{herrera2019reversals}, which spreads the entries across several tables whose headers vary; the loader locates each header row by its summary and discipline cells and reads the reversal identifier, the summary text, and the medical discipline, de-duplicating on the identifier. Each reversal is phrased as the pre-reversal belief and labeled false while retaining its apparent (pre-reversal) evidence grade, defaulting to high and using a per-item grade where the supplement provides one. This is the design element that places false claims at strong grades and thereby breaks the grade-truth correlation: all 395 reversals carry the pre-reversal era status and populate the otherwise-empty strong-grade-but-false cell, contributing 395 of the 396 reversed-high crossing-cell claims (the remaining one being a high-grade false CPSTF finding) and the single reversed-moderate claim.

\paragraph{Trialstreamer (19{,}870 claims).} The Trialstreamer database of automatically processed randomized controlled trial reports \cite{marshall2020trialstreamer} supplied graded claims at scale; its key-findings snippet is the claim text and its automated probability of low risk of bias is the quality signal. Because that probability is strongly right-skewed (most published trials score as low risk of bias), a fixed-threshold mapping collapses almost the entire source into a single grade and erases the grade axis; the claims were therefore graded by within-corpus quantile band, with the top quartile mapped to moderate, the middle two quartiles to low, and the bottom quartile to very low. This relative quality tier is reported as such, and a sensitivity analysis substituting an absolute risk-of-bias threshold for the quantile bands is reported alongside the main decodability result. Single-trial findings are not settled evidence, so every Trialstreamer claim is truth-uncertain; the source was read up to a cap of $20{,}000$ rows after a minimum-length filter, yielding a grade distribution of moderate~4{,}970, low~9{,}945, and very~low~4{,}955.

\paragraph{Evidence Inference (19{,}670 claims).} The Evidence Inference corpus \cite{lehman2019inferring} reports the outcome direction of individual trials (a significant increase, a significant decrease, or no significant difference for an intervention-comparator-outcome triple). Each significant-direction finding becomes a true directional claim and additionally emits its opposite direction as a verified false claim, a hard negative directly contradicted by the same trial, so the source carries a large truth-balanced true-and-false signal within one writing style; null findings are labeled truth-uncertain. Claim texts that appear as true under one trial and false under another trial's counterfactual were removed from both sides, so the source never labels a claim both ways. Evidence Inference carries no evidence grade. Its truth distribution is true~6{,}962, false~6{,}962, and uncertain~5{,}746 (Supplementary Table~\ref{stab:sourcetruth}); the matched 6{,}962 true and 6{,}962 false claims are what allow a truth direction to be learned in isolation from writing style, and they are the basis of the balanced grade-by-truth analysis.

\paragraph{EBM-NLP (4{,}853 claims).} The EBM-NLP corpus of PICO-annotated trial abstracts \cite{nye2018ebmnlp} contributes claim-text and topic diversity only; it carries neither a grade nor a truth label and its claims are truth-uncertain and ungraded. Each abstract yields one templated claim (``$\langle$intervention$\rangle$ affects $\langle$outcome$\rangle$ in $\langle$population$\rangle$'') built from the longest contiguous annotated span for each PICO element, capped at twelve words per span.

\subsection*{Assembly, filtering, and splitting}

After concatenation, exact duplicate claim texts were removed across the whole corpus. Three surface covariates were then computed for every claim: claim length in words, a hedge-term density measured against a fixed hedging lexicon, and a token-frequency proxy. A degenerate-claim filter removed any row shorter than four words, any row less than half alphabetic by character (which catches statistics fragments and bare citations), and any erratum, author-correction, retraction, or withdrawal notice, identified by a fixed pattern. The surviving claims number $45{,}134$, of which $20{,}611$ carry a four-level grade; the per-source claim counts after filtering are trialstreamer~19{,}870, evidence\_inference~19{,}670, ebm\_nlp~4{,}853, reversal~395, cpstf~224, and uspstf~122 (Supplementary Table~\ref{stab:sourcepipeline}). The corpus is $15.9\%$ true ($7{,}188$), $16.3\%$ false ($7{,}378$), and $67.7\%$ truth-uncertain ($30{,}568$), and $45.7\%$ graded.

The grade-by-source and grade-by-truth structure of the assembled corpus is what motivates the controls in the main analyses. Grade is strongly confounded with source: very-low claims are almost entirely from Trialstreamer ($4{,}955$ of $5{,}023$), and within the high grade every true claim is a CPSTF finding while every false claim is a reversal (Supplementary Tables~\ref{stab:gradesource} and the grade-by-truth contingency in the main text). Grade is also associated with truth: the high grade is $396$ false against $117$ true, whereas the low and very-low grades are predominantly truth-uncertain ($9{,}995$ of $10{,}002$ and $5{,}004$ of $5{,}023$ respectively). A small set of crossing cells decouples the two, namely $396$ reversed-high, $117$ settled-true, $102$ contested-true, $19$ settled-false, $7$ weak-true, and $1$ reversed-moderate claims.

Claims were partitioned at the claim level into training ($31{,}592$), validation ($6{,}771$), and test ($6{,}771$) sets by a $70/15/15$ split stratified on a composite key: each rare crossing cell forms its own stratum, with a floor of two held-out claims per cell so that the settled-false, weak-true, and reversed cells appear in both validation and test, while every remaining claim is stratified by grade so the bulk of the corpus stays grade-balanced across splits. The single reversed-moderate claim necessarily falls in one partition. The graded subset splits into $14{,}425$ training, $3{,}093$ validation, and $3{,}093$ test claims, and per grade the test split holds $77$ high, $1{,}501$ low, $761$ moderate, and $754$ very-low claims, the denominators for all held-out grade metrics.

\subsection*{Balanced grade-by-truth grid}

The separability analysis is run on a separate analysis set, not on the corpus above, because in the corpus the low and very-low grades carry essentially no truth labels and the only populated truth cells sit at the high and moderate grades, so grade and truth never vary independently. To build a set on which they do, the claims that already carry both a grade and a truth label were combined with Evidence Inference claims, which carry a trial-direction truth label but no grade and were each assigned a heuristic grade that is monotone in trial strength (a single significant trial mapped to low certainty, a null finding to very low); this heuristic is stated as such and used only to populate the otherwise-empty weak-grade truth cells. Within each grade the two truth cells were equalized to their shared size, capped at $2{,}000$ claims per cell, and a grade was retained only if both of its truth cells reached at least $50$ claims. On the assembled data this retained the low grade ($2{,}000$ claims per truth cell) and the high grade ($117$ per cell) and dropped the moderate and very-low grades, yielding a grid on which grade and truth are essentially uncorrelated. Because the truth direction must not become a source detector, it was learned on Evidence Inference alone, a within-source, single-template, truth-balanced set. All Evidence Inference claims were already present in the corpus, so no additional model inference was required to assemble the grid.

\subsection*{Worked examples}

To make the claim object concrete, Supplementary Table~\ref{stab:examples} gives a representative claim from each source with its harmonized grade and truth label, and Supplementary Table~\ref{stab:crossing} gives one claim from each of the six crossing cells, the rare claims for which grade and truth are decoupled. The claims from USPSTF, CPSTF, and the reversals are produced verbatim by the pipeline's own templates and are shown as written; the trial-text examples from Trialstreamer and Evidence Inference are lightly edited and marked as illustrative, since their wording derives from the underlying trial reports.

Two mechanisms are easier to see worked out than described. The Evidence Inference counterfactual, which supplies the matched true and false claims that decouple truth from writing style, is shown in Box~1: a reported trial direction becomes a true claim, and its exact opposite becomes a verified false claim that differs only in the asserted direction. A medical reversal, which places a false claim at a strong evidence grade and is the longest claim type in the corpus, is shown verbatim in Box~2.

\begin{claimbox}
\textbf{Box 1. Evidence Inference: a reported trial direction and its generated counterfactual (illustrative).}\\[3pt]
\textit{Trial.} A randomized trial compared vitamin~A supplementation against placebo and reported a significant increase in serum iron.\\[3pt]
\textit{True claim (reported direction).} ``Vitamin~A significantly increases serum iron compared with placebo.'' \quad[\,grade = N/A, truth = true\,]\\[3pt]
\textit{Counterfactual (opposite direction, a verified hard negative).} ``Vitamin~A significantly decreases serum iron compared with placebo.'' \quad[\,grade = N/A, truth = false\,]\\[3pt]
\textit{Why this matters.} The two claims share intervention, comparator, outcome, and writing style and differ only in the asserted direction, so an estimator that tells them apart must read the trial's finding and not surface wording. Across the corpus, this construction yields $6{,}962$ matched true and $6{,}962$ false claims, the within-style truth signal used to learn a truth direction and to build the balanced grade-by-truth grid. A null trial result has no clean opposite and is labeled truth-uncertain instead.
\end{claimbox}

\clearpage
\begin{claimbox}
\textbf{Box 2. A medical reversal: a false claim at a strong evidence grade (verbatim).}\\[3pt]
\textit{Source.} eLife medical-reversals supplement \cite{herrera2019reversals}. \textit{Grade} = high (apparent, pre-reversal). \textit{Truth} = false. \textit{Crossing cell} = reversed\_high. \textit{Era} = pre-reversal.\\[3pt]
\textit{Claim (verbatim).} ``In rural Asia, pesticide self-poisoning is a common health issue. Restricting access to pesticides has been previously shown to reduce both method-specific and all-cause suicide rates in certain Asian countries. The WHO advocates the use of locked boxes for storing pesticides in farming areas. In this study, lockable storage containers for pesticides were compared to usual practice in households (27{,}091 households to lockable storage containers and 26{,}291 households without) in rural areas of Sri Lanka. Lockable storage containers did not reduce the risk of pesticide self-poisoning (293/100{,}000 vs.\ 318/100{,}000 for intervention and control groups, respectively; $p=0.33$). This is a reversal of lockable storage containers to reduce self-poisoning.''\\[3pt]
\textit{Why this matters.} The claim states the pre-reversal belief that a later trial overturned, so it reads as a confident, strong-evidence recommendation yet is factually false. Placing such claims at the high grade is what populates the otherwise-empty strong-grade-but-false cell and lets the analysis separate evidence grade from factual truth. Reversals are also the longest claims in the corpus (median 130 words), in contrast to the nine-word USPSTF claims, which illustrates the lexical heterogeneity across sources.
\end{claimbox}

\begin{table}[t]
\centering
\caption{Per-source summary of the assembled clinical claim corpus ($n=45{,}134$ claims). For each of the six public sources, the table gives the number of claims after deduplication and degenerate-claim filtering, whether the source carries a harmonized evidence grade, the grading signal used, the factual-truth labels it contributes, and the median claim length in words. N/A, not applicable.}
\label{stab:sourcepipeline}
\footnotesize
\setlength{\tabcolsep}{5pt}
\renewcommand{\arraystretch}{1.2}
\begin{tabular}{@{}lr p{0.12\columnwidth} p{0.20\columnwidth} p{0.18\columnwidth} r@{}}
\toprule
Source & Claims & Graded & Grading signal & Truth labels & Median words \\
\midrule
Trialstreamer & 19{,}870 & yes & RoB quantile band & uncertain only & 22 \\
Evidence Inference & 19{,}670 & no & N/A & true / false / uncertain & 15 \\
EBM-NLP & 4{,}853 & no & N/A & uncertain only & 24 \\
Reversals & 395 & yes & apparent pre-reversal grade & false only & 130 \\
CPSTF & 224 & yes & finding direction and strength & true / false / uncertain & 15 \\
USPSTF & 122 & yes & letter grade A--I & true / false / uncertain & 9 \\
\midrule
Total & 45{,}134 & & & & 20 \\
\bottomrule
\end{tabular}
\end{table}

\begin{table}[t]
\centering
\caption{Grade-by-source composition of the graded subset ($n=20{,}611$ claims). Each cell is the number of graded claims of the given harmonized grade contributed by the given source; a dash marks a grade a source does not contribute. Evidence Inference and EBM-NLP carry no grade and are omitted. USPSTF is the only source spanning all four grades.}
\label{stab:gradesource}
\footnotesize
\setlength{\tabcolsep}{8pt}
\renewcommand{\arraystretch}{1.15}
\begin{tabular}{@{}lrrrr@{}}
\toprule
Harmonized grade & Trialstreamer & Reversals & CPSTF & USPSTF \\
\midrule
high & -- & 395 & 106 & 12 \\
moderate & 4{,}970 & -- & 68 & 35 \\
low & 9{,}945 & -- & 50 & 7 \\
very low & 4{,}955 & -- & -- & 68 \\
\midrule
Total & 19{,}870 & 395 & 224 & 122 \\
\bottomrule
\end{tabular}
\end{table}

\begin{table}[t]
\centering
\caption{Factual-truth composition by source across the full corpus ($n=45{,}134$ claims). Each cell is the number of claims with the given truth label from the given source. Truth labels are assigned independently of grade. Evidence Inference contributes the matched $6{,}962$ true and $6{,}962$ false claims that decouple truth from writing style.}
\label{stab:sourcetruth}
\footnotesize
\setlength{\tabcolsep}{8pt}
\renewcommand{\arraystretch}{1.15}
\begin{tabular}{@{}lrrrr@{}}
\toprule
Source & True & False & Uncertain & Total \\
\midrule
Trialstreamer & 0 & 0 & 19{,}870 & 19{,}870 \\
Evidence Inference & 6{,}962 & 6{,}962 & 5{,}746 & 19{,}670 \\
EBM-NLP & 0 & 0 & 4{,}853 & 4{,}853 \\
Reversals & 0 & 395 & 0 & 395 \\
CPSTF & 172 & 2 & 50 & 224 \\
USPSTF & 54 & 19 & 49 & 122 \\
\midrule
Total & 7{,}188 & 7{,}378 & 30{,}568 & 45{,}134 \\
\bottomrule
\end{tabular}
\end{table}

\begin{table}[p]
\centering
\caption{Composition of the clinical evidence-grade claim corpus. Counts are numbers of claims. The graded subset comprises the claims carrying a four-level evidence grade; truth labels are assigned independently of grade, and the crossing cells are the rare claims for which grade and truth are decoupled (for example, settled-false and weak-but-true claims). Splits are reported as all claims and the graded subset.}
\label{stab:corpus}
\small
\setlength{\tabcolsep}{10pt}
\renewcommand{\arraystretch}{1.1}
\begin{tabular}{@{}lr@{\hspace{6em}}lr@{}}
\toprule
Category & Count & Category & Count \\
\midrule
\multicolumn{2}{@{}l}{\textit{Sources}} &
\multicolumn{2}{l@{}}{\textit{Truth labels}} \\
trialstreamer & 19{,}870 & true & 7{,}188 \\
evidence\_inference & 19{,}670 & false & 7{,}378 \\
ebm\_nlp & 4{,}853 & uncertain & 30{,}568 \\
reversal & 395 & & \\
cpstf & 224 &
\multicolumn{2}{l@{}}{\textit{Crossing cells (grade-truth decoupled)}} \\
uspstf & 122 & reversed\_high & 396 \\
Total & 45{,}134 & settled\_true & 117 \\
\midrule
\multicolumn{2}{@{}l}{\textit{Evidence grades (graded subset)}} &
contested\_true & 102 \\
high & 513 & settled\_false & 19 \\
moderate & 5{,}073 & weak\_true & 7 \\
low & 10{,}002 & reversed\_moderate & 1 \\
very\_low & 5{,}023 & & \\
Total graded & 20{,}611 &
\multicolumn{2}{l@{}}{\textit{Splits (all / graded)}} \\
& & train & 31{,}592 / 14{,}425 \\
& & val & 6{,}771 / 3{,}093 \\
& & test & 6{,}771 / 3{,}093 \\
\bottomrule
\end{tabular}
\end{table}

\begin{table}[t]
\centering
\caption{Harmonization of native framework grades to the common four-level evidence scale. Each entry is the native grade or finding type that maps to the harmonized grade in that row; ``N/A'' marks grades a source does not produce. Factual-truth labels were assigned independently of grade: a claim is true for USPSTF grades A, B, and C and for CPSTF recommendations, false for medical reversals, USPSTF grade D, and CPSTF recommended-against findings, and truth-uncertain for USPSTF grade I, insufficient-evidence findings, and single-trial or null trial results. A sensitivity analysis substitutes an absolute risk-of-bias threshold for the Trialstreamer quantile bands. N/A, not applicable.}
\label{stab:harmonization}
\footnotesize
\setlength{\tabcolsep}{7pt}
\renewcommand{\arraystretch}{1.2}
\begin{tabular}{@{}ll p{0.30\columnwidth} ll@{}}
\toprule
Harmonized grade & USPSTF & CPSTF & Trialstreamer & Reversals \\
\midrule
high & A & Recommended / Against (strong) & N/A & apparent high \\
moderate & B & Recommended (sufficient); Against (no strength) & top band & apparent moderate \\
low & C & Insufficient evidence & middle band & N/A \\
very low & D, I & N/A & bottom band & N/A \\
\bottomrule
\end{tabular}
\end{table}

\begin{table}[t]
\centering
\caption{The 22 open-weight language models tested in the study, grouped by family. Parameters are in billions. The peak layer is the transformer layer at which the grade estimator was most decodable on the validation split (layer 0 is the embedding output), and the depth fraction is the peak layer as a fraction of the model's total layers. HuatuoGPT-o1 is a medical reasoning model and is grouped with the reasoning-distilled family.}
\label{stab:models}
\footnotesize
\setlength{\tabcolsep}{5pt}
\renewcommand{\arraystretch}{1.15}
\begin{tabular}{@{}llrrr@{}}
\toprule
Model & Hugging Face checkpoint & Params (B) & Peak layer & Depth \\
\midrule
\multicolumn{5}{@{}l}{\textit{General-domain}} \\
Qwen3-0.6B & \texttt{Qwen/Qwen3-0.6B} & 0.6 & 27 & 0.96 \\
Qwen3-1.7B & \texttt{Qwen/Qwen3-1.7B} & 1.7 & 28 & 1.00 \\
Qwen3-4B & \texttt{Qwen/Qwen3-4B} & 4 & 13 & 0.36 \\
Qwen3-8B & \texttt{Qwen/Qwen3-8B} & 8 & 20 & 0.56 \\
Qwen3-14B & \texttt{Qwen/Qwen3-14B} & 14 & 37 & 0.93 \\
Qwen3-32B & \texttt{Qwen/Qwen3-32B} & 32 & 55 & 0.86 \\
Phi-3.5-mini & \texttt{microsoft/Phi-3.5-mini-instruct} & 3.8 & 20 & 0.63 \\
Mistral-7B-v0.3 & \texttt{mistralai/Mistral-7B-Instruct-v0.3} & 7 & 16 & 0.50 \\
Llama-3.1-8B & \texttt{meta-llama/Llama-3.1-8B-Instruct} & 8 & 28 & 0.88 \\
Gemma-2-9B & \texttt{google/gemma-2-9b-it} & 9 & 27 & 0.64 \\
Gemma-2-27B & \texttt{google/gemma-2-27b-it} & 27 & 45 & 0.98 \\
Llama-3.3-70B & \texttt{meta-llama/Llama-3.3-70B-Instruct} & 70 & 58 & 0.73 \\
\midrule
\multicolumn{5}{@{}l}{\textit{Medical domain-adapted}} \\
BioMistral-7B & \texttt{BioMistral/BioMistral-7B} & 7 & 31 & 0.97 \\
Meditron-7B & \texttt{epfl-llm/meditron-7b} & 7 & 31 & 0.97 \\
OpenBioLLM-8B & \texttt{aaditya/OpenBioLLM-Llama3-8B} & 8 & 20 & 0.63 \\
Med42-8B & \texttt{m42-health/Llama3-Med42-8B} & 8 & 29 & 0.91 \\
MedGemma-27B & \texttt{google/medgemma-27b-it} & 27 & 34 & 0.55 \\
Med42-70B & \texttt{m42-health/Llama3-Med42-70B} & 70 & 76 & 0.95 \\
\midrule
\multicolumn{5}{@{}l}{\textit{Reasoning-distilled}} \\
R1-Distill-Qwen-7B & \texttt{deepseek-ai/DeepSeek-R1-Distill-Qwen-7B} & 7 & 22 & 0.79 \\
R1-Distill-Llama-8B & \texttt{deepseek-ai/DeepSeek-R1-Distill-Llama-8B} & 8 & 22 & 0.69 \\
R1-Distill-Qwen-32B & \texttt{deepseek-ai/DeepSeek-R1-Distill-Qwen-32B} & 32 & 64 & 1.00 \\
HuatuoGPT-o1-8B & \texttt{FreedomIntelligence/HuatuoGPT-o1-8B} & 8 & 28 & 0.88 \\
\bottomrule
\end{tabular}
\end{table}

\begin{table}[t]
\centering
\caption{One representative clinical claim from each of the six sources, with its harmonized evidence grade and factual-truth label. USPSTF, CPSTF, and reversal claims are pipeline-generated and shown verbatim (the reversal is truncated here; its full text appears in Box~2); Trialstreamer and Evidence Inference claims are lightly edited and shown as illustrative. N/A, not applicable (the source carries no harmonized grade).}
\label{stab:examples}
\footnotesize
\setlength{\tabcolsep}{7pt}
\renewcommand{\arraystretch}{1.25}
\begin{tabular}{@{}l p{0.45\columnwidth} ll@{}}
\toprule
Source & Example claim & Grade & Truth \\
\midrule
USPSTF & Screening for syphilis infection during pregnancy & high & true \\
CPSTF & School-based self-management interventions for children and adolescents with asthma is an effective community health intervention for asthma & high & true \\
Trialstreamer & Polished surfaces showed insignificantly higher surface roughness than glazed surfaces & very low & uncertain \\
Evidence Inference & Vitamin~A significantly increases serum iron compared with placebo & N/A & true \\
EBM-NLP & Inhaled corticosteroid affects exacerbation rate in adults with asthma & N/A & uncertain \\
Reversals & Lockable pesticide storage containers did not reduce the risk of pesticide self-poisoning \ldots\ (full text in Box~2) & high & false \\
\bottomrule
\end{tabular}
\end{table}

\begin{table}[t]
\centering
\caption{One claim from each of the six crossing cells, the rare claims for which evidence grade and factual truth are decoupled, with the source and the contributing cell count. These cells are the basis of the balanced grade-by-truth analysis. Claims are pipeline-generated and shown verbatim, truncated where long.}
\label{stab:crossing}
\footnotesize
\setlength{\tabcolsep}{7pt}
\renewcommand{\arraystretch}{1.25}
\begin{tabular}{@{}l p{0.40\columnwidth} lll@{}}
\toprule
Crossing cell & Example claim & Grade & Truth & $n$ \\
\midrule
reversed\_high & Lockable pesticide storage containers did not reduce pesticide self-poisoning & high & false & 396 \\
settled\_true & Screening for syphilis infection during pregnancy & high & true & 117 \\
contested\_true & Screening for intimate partner violence in women of reproductive age & moderate & true & 102 \\
settled\_false & Screening for genital herpes infection & very low & false & 19 \\
weak\_true & Statin use for the primary prevention of cardiovascular disease in adults aged 40 to 75 with risk factors & low & true & 7 \\
reversed\_moderate & Policies facilitating the transfer of juveniles to adult justice systems is an effective community health intervention for violence prevention & moderate & false & 1 \\
\bottomrule
\end{tabular}
\end{table}

%%%%%%%%%%%%
\begin{figure*}[p]
\centering
\includegraphics[width=\textwidth]{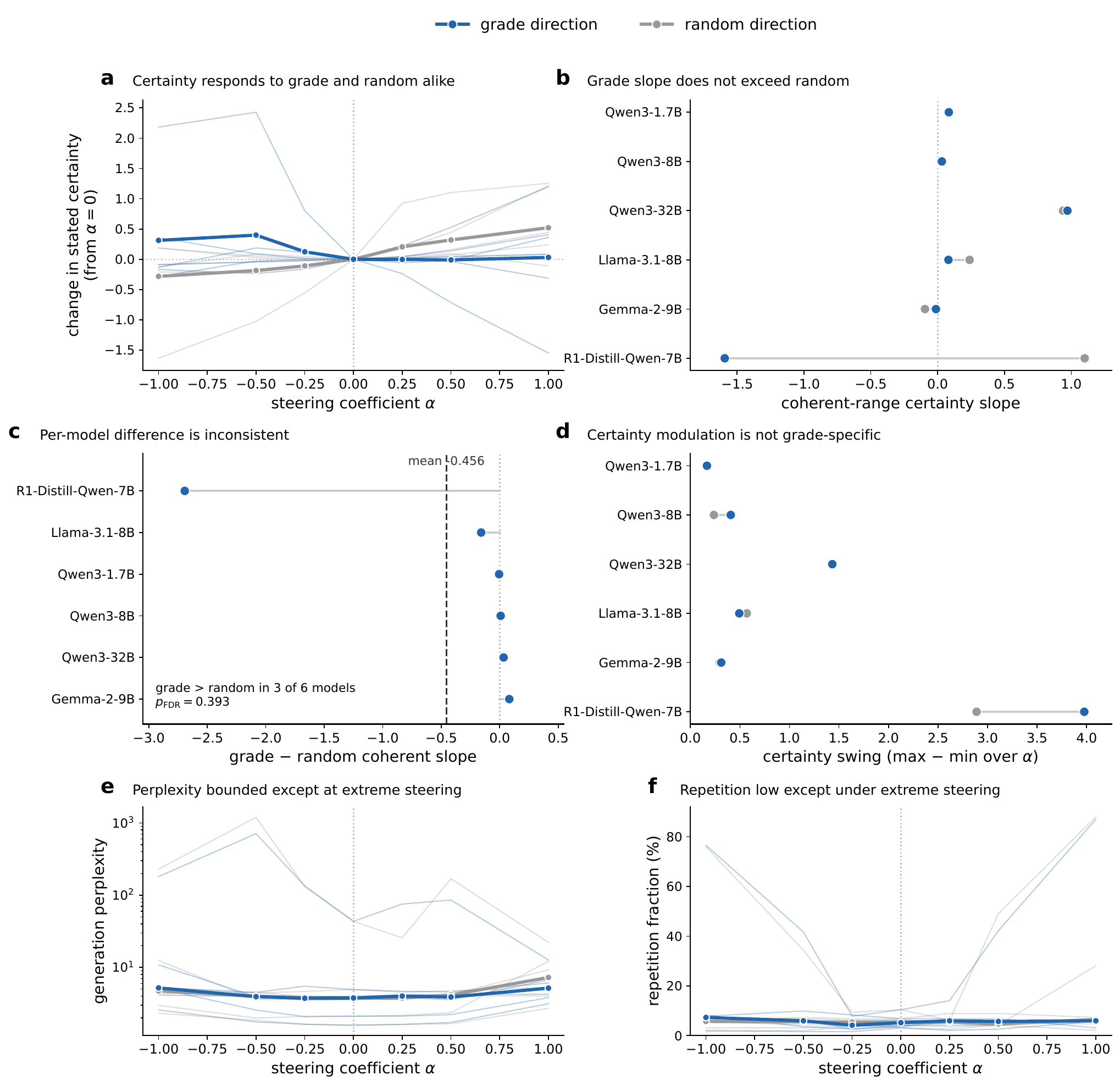}
\caption{Activation steering along the evidence-grade direction does not control what the models express, for six of the seven models selected for steering (MedGemma-27B is excluded because its decoder layers could not be located for injection). At each model's peak layer the unit-norm ridge grade direction (grade) or a matched random direction (random) was added to the residual stream during generation, scaled by the mean residual norm, with a signed coefficient $\alpha$ swept from $-1$ to $1$, and the model's stated certainty was read from its certainty-token probabilities; the random direction is averaged over three random draws. \textbf{a}, Change in stated certainty from its $\alpha=0$ value against $\alpha$, centered per model so the dose-response is comparable across models, with one thin line per model and the cross-model mean in bold, for grade and random steering. \textbf{b}, Per-model slope of stated certainty against $\alpha$ over the coherent range, for grade and random steering, as a paired marker per model with a zero reference. \textbf{c}, Per-model grade-minus-random coherent slope, sorted, with a zero reference and the cross-model mean ($-0.456$); the grade slope exceeds the random slope in three of the six models and the mean difference is not significant ($p_{\mathrm{FDR}}=0.393$). \textbf{d}, Per-model maximum-minus-minimum stated certainty across the swept $\alpha$ range, for grade and random steering, as a paired marker per model. \textbf{e}, Generation perplexity of the steered continuations against $\alpha$ on a logarithmic axis, with one thin line per model and the cross-model median in bold, for grade and random steering, as a coherence control. \textbf{f}, Repetition fraction of the steered continuations against $\alpha$, for grade and random steering, as a coherence control. Grade-direction quantities are shown in blue and random-direction quantities in gray throughout.}
\label{sfig:steering}
\end{figure*}
%%%%%%%%%%%%

\end{document}